\documentclass{article}

\usepackage{microtype}
\usepackage{graphicx}
\usepackage{subfigure}
\usepackage{booktabs} 
\usepackage{floatrow}

\usepackage{hyperref}
\usepackage[accepted]{icml2026}
\usepackage[utf8]{inputenc} 
\usepackage[T1]{fontenc}    
\usepackage[most]{tcolorbox}
\usepackage{adjustbox}
\usepackage[table]{xcolor}
\usepackage{xspace}
\newcolumntype{L}[1]{>{\raggedright\let\newline\\\arraybackslash\hspace{0pt}}m{#1}}
\newcolumntype{C}[1]{>{\centering\let\newline\\\arraybackslash\hspace{0pt}}m{#1}}
\newcolumntype{R}[1]{>{\raggedleft\let\newline\\\arraybackslash\hspace{0pt}}m{#1}}

\newcommand{\ignore}[1]{}

\DeclareMathAlphabet{\mathbfit}{OML}{cmm}{b}{it}

\makeatletter
\DeclareRobustCommand\onedot{\futurelet\@let@token\@onedot}
\def\@onedot{\ifx\@let@token.\else.\null\fi\xspace}

\makeatother

\definecolor{MyDarkBlue}{rgb}{0,0.08,1}
\definecolor{MyAqua}{rgb}{0,0.7,0.7}
\definecolor{MyDarkGreen}{rgb}{0.02,0.6,0.02}
\definecolor{MyDarkRed}{rgb}{0.8,0.02,0.02}
\definecolor{MyDarkOrange}{rgb}{0.40,0.2,0.02}
\definecolor{MyPurple}{RGB}{111,0,255}
\definecolor{MyRed}{rgb}{1.0,0.0,0.0}
\definecolor{MyGold}{rgb}{0.75,0.6,0.12}
\definecolor{MyDarkgray}{rgb}{0.66, 0.66, 0.66}

\definecolor{LightCyan}{rgb}{0.88,1,1}

\definecolor{ourmethod}{gray}{0.93}

\definecolor{myblue2}{rgb}{0,0.08,0.45}
\usepackage{multirow}
\definecolor{panelA}{HTML}{FFF6DC} 
\definecolor{panelB}{HTML}{EEF7F1} 
\newtcolorbox{empheqboxed}{colback=Gray!20, 
 colframe=white,
 width=\textwidth,
 sharpish corners,
 top=1mm, 
 bottom=0pt,
 left=2pt,
 right=2pt
}
\definecolor{lightyellow}{HTML}{FFFFED}
\definecolor{lightpink}{HTML}{F8F1EF}

\usepackage{hyperref}       
\usepackage{url}            
\usepackage{booktabs}       
\usepackage{minitoc}       
\usepackage{amsfonts}       
\usepackage{nicefrac}       
\usepackage{microtype}      
\usepackage[dvipsnames]{xcolor}         

\usepackage{amsmath,amsfonts,bm}

\def\eqref#1{equation~\ref{#1}}

\def\1{\bm{1}}

\def\rv{{\textnormal{v}}}

\def\rva{{\mathbf{a}}}

\def\rvg{{\mathbf{g}}}

\def\rvo{{\mathbf{o}}}

\def\rvs{{\mathbf{s}}}

\def\rvx{{\mathbf{x}}}

\def\rvz{{\mathbf{z}}}

\def\vm{{\bm{m}}}

\DeclareMathAlphabet{\mathsfit}{\encodingdefault}{\sfdefault}{m}{sl}
\SetMathAlphabet{\mathsfit}{bold}{\encodingdefault}{\sfdefault}{bx}{n}

\usepackage{url}
\usepackage{lipsum}
\usepackage{pifont}
\usepackage{etoolbox}
\usepackage{lineno} 
\usepackage{microtype}
\usepackage{graphicx}
\usepackage{booktabs} 
\usepackage{floatrow}
\usepackage{hyperref}
\usepackage{tikz-cd}
\usepackage{amssymb, amsmath, amsthm, amsfonts}
\usepackage{soul}
\usepackage{tikz}
\usepackage{wrapfig}
\usepackage{algpseudocode}
\usepackage[utf8]{inputenc} 
\usepackage[T1]{fontenc}    
\usepackage{hyperref}       
\usepackage{url}            
\usepackage{nicefrac}       
\usepackage{bbm}
\usepackage{xcolor}
\usepackage{color}
\usepackage{threeparttable}
\usepackage{caption}
\usepackage{subcaption}
\usepackage{multibib}
\definecolor{lightbeige}{RGB}{250,243,224}

\usepackage[most]{tcolorbox}
\tcbset{
  beigenote/.style={
    enhanced,
    breakable,               
    width=\linewidth,        
    colback=lightbeige,      
    colframe=lightbeige!60!black, 
    boxrule=0.5pt,          
    arc=2pt,                
    left=8pt,right=8pt,top=6pt,bottom=6pt, 
  }
}
\newtcolorbox{beigebox}[1][]{beigenote,#1}

\newtheorem{Definition}{Definition}

\usepackage{thmtools,thm-restate}

\definecolor{candypink}{rgb}{0.89, 0.44, 0.48}          %
\definecolor{mediumaquamarine}{rgb}{0.4, 0.8, 0.67}     %
\definecolor{azure}{rgb}{0.0, 0.5, 1.0}                 %
\definecolor{awesome}{rgb}{1.0, 0.13, 0.32}             %
\definecolor{myred}{HTML}{F54254}
\definecolor{myblue}{HTML}{598BE7}
\definecolor{mydarkblue}{HTML}{385492}

\definecolor{c1}{RGB}{255,75,0}
\definecolor{c2}{RGB}{0,180,255}
\definecolor{LightCyan}{rgb}{0.88,1,1}
\definecolor{Gray}{gray}{0.9}
\usepackage{diagbox}
\usepackage[
    separate-uncertainty = true,
    multi-part-units = repeat
]{siunitx}
\usepackage{microtype}
\usepackage{booktabs} 
\usepackage{multirow}
\usepackage{pifont}
\usepackage{colortbl}

\usepackage{paralist,tabularx}
\usepackage{array}
\newcolumntype{?}{!{\vrule width 1pt}}
\usetikzlibrary{fit,positioning}

\definecolor{customRed}{RGB}{190,110,113}
\colorlet{mylinkcolor}{violet}
\definecolor{darkred}{rgb}{0.55, 0.0, 0.0}

\colorlet{mycitecolor}{RoyalBlue}
\colorlet{myurlcolor}{RoyalPurple}
\hypersetup{
  citecolor  = myblue2,
  linkcolor = darkred,
  urlcolor = darkred,
  colorlinks = true
}

\newcommand{\pa}[1]{\text{pa}(#1)}

\newcommand{\mist}{\texttt{MIST-WM}\xspace}
\definecolor{myredcolor}{RGB}{215,48,39}
\definecolor{mygreencolor}{RGB}{26,152,80}
\definecolor{goldcolor}{RGB}{255,165,0}

\makeatletter
\setbox0\hbox{$\xdef\scriptratio{\strip@pt\dimexpr
\numexpr(\sf@size*65536)/\f@size sp}$}
\newcommand{\scriptveryshortarrow}[1][3pt]{{%
    \vcenter{\hbox{\rule[\scriptratio\dimexpr-.2pt\relax]
    {\scriptratio\dimexpr#1\relax}{\scriptratio\dimexpr.4pt\relax}}}%
    \mkern-4mu\hbox{\let\f@size\sf@size\usefont{U}{lasy}{m}{n}\symbol{41}}}}
\usepackage{pifont}

\usepackage{wrapfig}

\usepackage[utf8]{inputenc} 
\usepackage[T1]{fontenc}    
\usepackage{courier}
\usepackage{hyperref}       
\usepackage{url}            

\usepackage{booktabs}       
\usepackage{amsfonts}       
\usepackage{amsmath} 
\usepackage{nicefrac}       
\usepackage{wrapfig}
\usepackage{multirow}

\usepackage{tikz}
\usetikzlibrary{fit,matrix,positioning, 
decorations.pathreplacing,calc,decorations,arrows,shadows,patterns}
\usetikzlibrary{arrows,bayesnet,fit,positioning,shapes,matrix}
\newdimen\arrowsize
\pgfarrowsdeclare{arcsq}{arcsq}
{
	\arrowsize=0.2pt
	\advance\arrowsize by .5\pgflinewidth
	\pgfarrowsleftextend{-4\arrowsize-.5\pgflinewidth}
	\pgfarrowsrightextend{.5\pgflinewidth}
}
{
	\arrowsize=0.8pt
	\advance\arrowsize by .5\pgflinewidth
	\pgfsetdash{}{0pt} 
	\pgfsetroundjoin   
	\pgfsetroundcap    
	\pgfpathmoveto{\pgfpoint{0\arrowsize}{0\arrowsize}}
	\pgfpatharc{-90}{-140}{4\arrowsize}
	\pgfusepathqstroke
	\pgfpathmoveto{\pgfpointorigin}
	\pgfpatharc{90}{140}{4\arrowsize}
	\pgfusepathqstroke
}

\usepackage{amsmath}
\usepackage{amssymb}
\usepackage{mathtools}
\usepackage{amsthm}
\usepackage{textcomp}
\usepackage{booktabs}
\usepackage{makecell}

\definecolor{myred1}{HTML}{D62727}

\usepackage{titletoc}

\usepackage{xcolor,colortbl}
\definecolor{ourmethod}{gray}{0.93}
\definecolor{myredcolor}{RGB}{215,48,39}
\definecolor{mygreencolor}{RGB}{26,152,80}
\usepackage{pifont}

\definecolor{dred}{rgb}{0.8, 0.0, 0.0}

\usepackage{enumitem}

\usepackage[utf8]{inputenc} 
\usepackage[T1]{fontenc}    
\usepackage{hyperref}       
\usepackage{url}            
\usepackage{booktabs}       
\usepackage{amsfonts}       
\usepackage{nicefrac}       
\usepackage{microtype}      
\usepackage{xcolor}         

\icmltitlerunning{Learning Task-Sufficient World Models by Synergizing Agentic Exploration and Structured Modeling}

\begin{document}

\twocolumn[
\icmltitle{Learning Task-Sufficient World Models by Synergizing Agentic Exploration and Structured Modeling}


\icmlsetsymbol{equal}{*}

\begin{icmlauthorlist}
\icmlauthor{Fan Feng}{equal,1,2}
\icmlauthor{Yujia Zheng}{equal,3}
\icmlauthor{Minghao Fu}{1}
\icmlauthor{Yongqiang Chen}{2,3}
\icmlauthor{Guangyi Chen}{2,3} \\
\icmlauthor{Kevin Murphy}{4}
\icmlauthor{Biwei Huang}{1}
\icmlauthor{Kun Zhang}{2,3}
\end{icmlauthorlist}

  \icmlaffiliation{1}{UCSD}
  \icmlaffiliation{2}{MBZUAI}
  \icmlaffiliation{3}{CMU}
  \icmlaffiliation{4}{UBC}

\icmlcorrespondingauthor{Fan Feng}{ffeng1017@gmail.com}

\icmlkeywords{Machine Learning, ICML}

\vskip 0.3in
]

\printAffiliationsAndNotice{\icmlEqualContribution} 

\newcommand{\fix}{\marginpar{FIX}}
\newcommand{\new}{\marginpar{NEW}}

\begin{abstract}
Learning and planning in imagination using world models provides an effective paradigm for training agents for decision-making. However, existing approaches often rely on high-dimensional pixel spaces or generic visual embeddings that retain many factors irrelevant to control, limiting efficiency and generalization across tasks. 
To this end, we study how agents can learn world models with representations that are \textit{task-specific}, \textit{minimal}, and \textit{sufficient} for decision making. We achieve this via a closed-loop synergy between the \textit{agent} and the \textit{world model}, in which structured world-model learning distills task-sufficient representations from informative interaction data. On the agent side, agents perform \textit{active probing} of the environment to collect informative trajectories that expose task-relevant latent factors, guided by an adaptive curriculum. On the world-model side, we learn \textit{structured representations} over observations to distill compact, task-sufficient latent states from the collected interaction data. This synergy enables the recovery of task-sufficient latent representations that capture all control-relevant factors. Leveraging these representations, the resulting policies achieve sample-efficient generalization in novel settings, including generalization across skills, object–skill compositions, and previously unseen tasks on continuous control and robotic manipulation benchmarks. Project website is  {\href{https://sites.google.com/view/mist-wm/home}{here}}.
\end{abstract}
\section{Introduction}
World models~\citep{ha2018world} are generative or predictive models that capture environment observation functions, dynamics, and rewards; in model-based RL (MBRL), it benefits planning, policy improvement, and simulated rollouts to reduce sample complexity~\citep{sutton1991dyna}. One way of learning the world model is to compress raw observations into meaningful latent states that support accurate prediction and control, facilitating reuse across tasks and improving generalization, e.g., Dreamer-series~\citep{Hafner2020Dream, hafner2021mastering, hafner2025mastering}, Efficient-Zero~\citep{schrittwieser2020mastering, ye2021mastering}, and TD-MPC~\citep{hansen2022temporal, hansen2024tdmpc}. Recent work also leverages pre-trained foundation models as strong perceptual encoders: using embeddings as the observation space, and the agent then learns dynamics on top of these embeddings via self-predictive manners~\citep{zhou2025dinowm, baldassarre2025back, kapoor2025pretrained, wang2025founder, assran2025v}.

These models often excel at learning visual dynamics and supporting downstream control. Yet their internal representations frequently entangle perceptual factors unrelated to control, which can yield brittle policies in complex or changing environments~\citep{wang2022denoised,liu2023learning, vafa2025has}. {By contrast, humans typically do not plan using pixel-level predictions or by tracking redundant perceptual details; instead, our planning relies on compact task-relevant representations~\citep{mastrogiuseppe2018linking, ho2022people, rajalingham2022recurrent, nayebi2023neural}.}
Motivated by this, we argue that learning world models with \emph{task-specific, minimal sufficient} latents, derived from raw observations or foundation-model priors, is essential for policy learning, as such representations preserve exactly the information needed for decision-making. 

In pursuit of these minimal and sufficient representations in world models, we draw inspiration from how humans acquire structured knowledge through interaction. A child actively probes the environment, selectively seeking informative experiences and gradually mastering abstract concepts in a structured manner \citep{schulz2008word, bonawitz2011double}.
Similarly, for agent learning with world models, we can decompose the problem into two coupled components: {interactive data curation} by the \textbf{agentic exploration} and a \textbf{structured world model} learning framework that distills task-sufficient representations. These two components form a closed-loop synergy centered on learning task-sufficient representations for world models. Agentic exploration determines \textit{what data are collected}, while structured world-model learning determines \textit{how task-relevant information is distilled} from that data. 
\begin{figure}[t]
\begin{center}
\centerline{\includegraphics[width=\columnwidth]{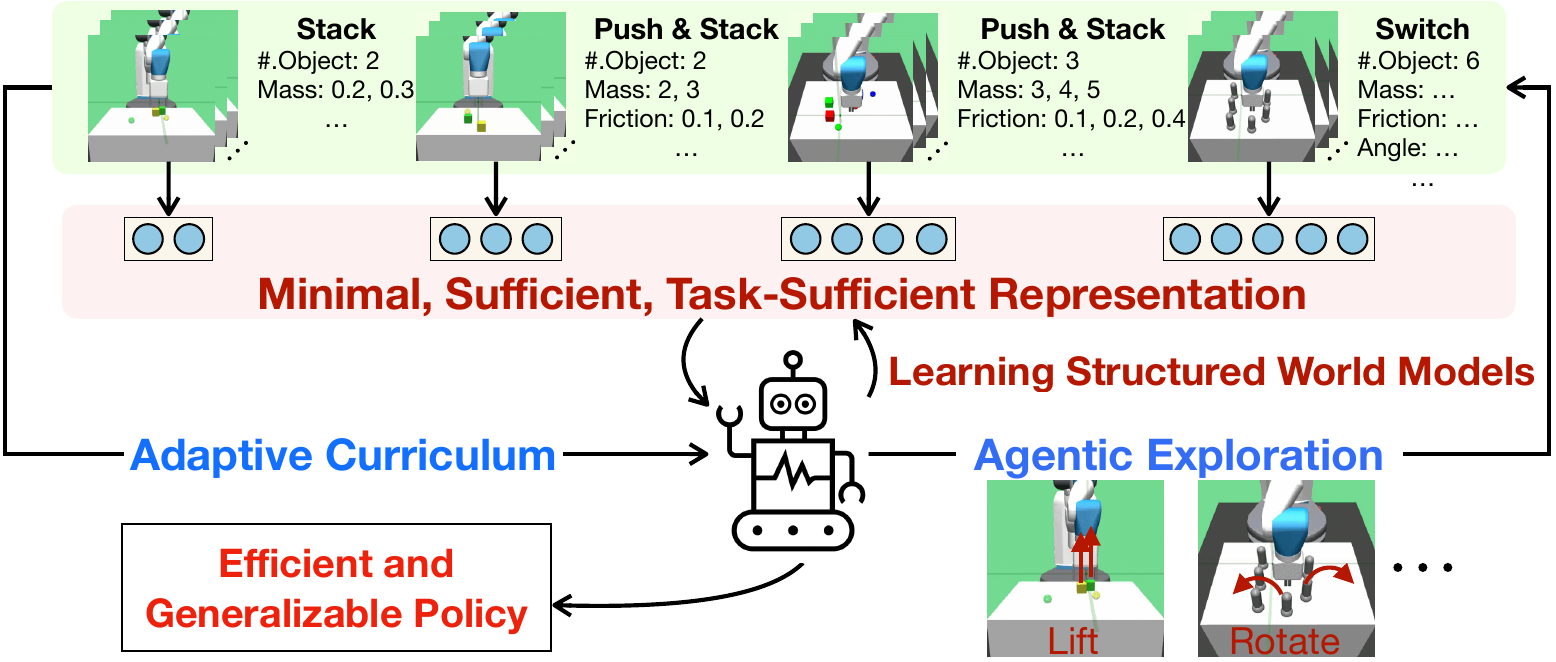}}
\caption{\textbf{Overview on} \mist: co-evolve on the agentic exploration and structured world models. The agent conducts active interventions to curate meaningful behavior data to identify the key latent factors in each new environment under an adaptive curriculum. With the curated data, a structured world model with minimal, sufficient, and task-specific representation is learned.}
\label{fig:overview}\vspace{-2mm}
\end{center}
\vspace{-5mm}
\end{figure}

We term this framework \mist (Minimal, Sufficient, and Task-specific World Model), which considers the general setting of agents facing a sequence of tasks, and co-evolves agentic exploration with structured world-model learning across these tasks (Fig.~\ref{fig:overview}). The agent learns world models sequentially, using each previously learned model as a prior for the next, yielding a latent space that progressively expands while remaining well-structured.
This is achieved through a closed-loop interaction between agentic exploration and world models. Upon encountering a new task, the agent actively designs and executes probing behaviors via self-supervised skills to collect trajectories that maximize expected information gain about task-specific latent factors. The interaction data is then used to learn a structured world model under objectives that enforce minimality and sufficiency, yielding compact and structured representations.
Together, these components enable \mist to (i) learn task-specific latent representations that are minimal and sufficient across a sequence of tasks, and (ii) support sample-efficient generalization in novel settings, as policies conditioned on sufficient representations can reuse control-relevant factors and adapt efficiently to novel tasks without irrelevant structure.

Overall, \mist offers an end-to-end recipe for what data to collect, how to learn the world model, and how to generalize across tasks via the reuse of learned representation. We evaluate on locomotion and robotic control benchmarks, in which \mist learns task-specific representations that are both minimal and sufficient, aligning with true system variables. With these compact representations, it achieves sample-efficient generalization in novel settings, at the skill level, in object–skill compositions, and on unseen tasks by composing the previously learned representation.



\begin{figure*}[t]
\begin{center}
\centerline{\includegraphics[width=.9\linewidth]{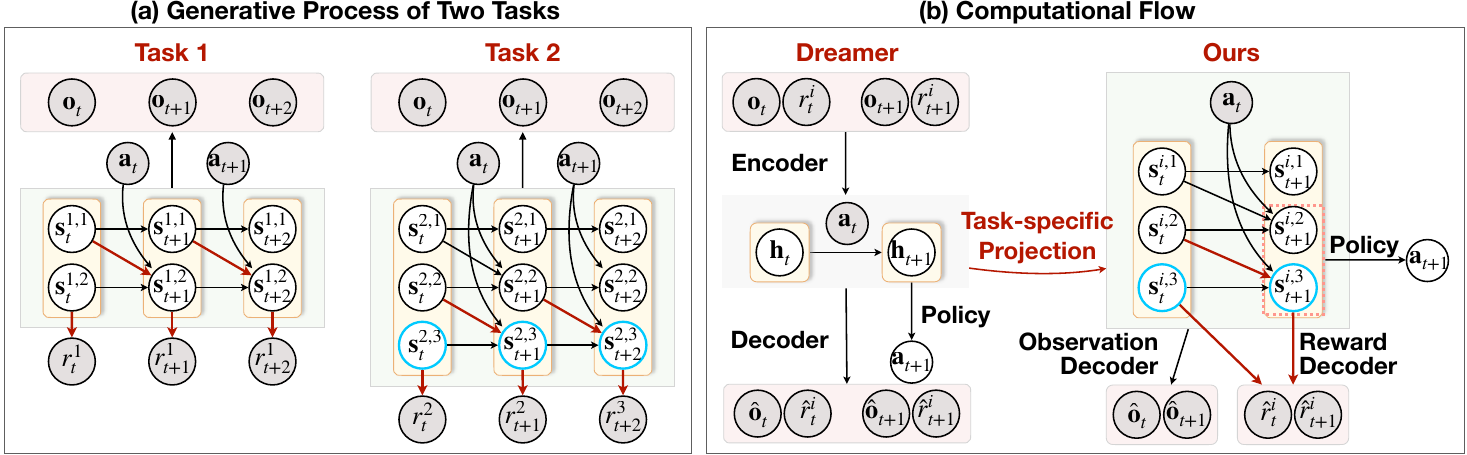}}
\caption{(a) Example DBNs that describe the generative process of two tasks. The grey and white nodes indicate observable and latent variables. The blue nodes are the newly expanded variables for Task 2. Red edges are the links from state variables to rewards. For simplicity, we omit the factorization of the state–observation mapping $\rvs \rightarrow \rvo$. (b) Computational flow of our proposed framework compared with the Dreamer-based world model.}
\label{fig:example}
\end{center}
\end{figure*}

\section{Preliminaries}
\label{sec:pre}
In this section, we formally define minimal and sufficient task-specific representations and the corresponding structured world models. We consider a general setting where an agent encounters a sequence of tasks whose underlying state factors may vary across tasks, e.g., new objects, dynamics, or reward-relevant variables may be introduced over time.

To capture the essential structures under this setting, we model the environment as a factored partially observable Markov decision process (POMDP)~\citep{boutilier2000stochastic} with an expandable latent state space, allowing new state variables to be introduced as the agent transitions to new tasks. We then define the \emph{Minimal, Sufficient, Task-specific} (MIST) states and outline the computational flow for identifying and using MIST within the world model.

\textbf{Factored POMDP with Changing Structure and Expandable States.}
We model the system as a factored POMDP, which can be represented as a tuple $\mathcal{M}=\big(\mathcal{S},\mathcal{A},\mathcal{O},\gamma,\mathcal{G},\mathbb{P}_s,\mathbb{P}_o,\mathcal{R},\rho_0\big)$,
where $\mathcal{S}$ is the (latent) state space, $\mathcal{A}$ the action space, $\mathcal{O}$ the observation space,
$\gamma\in(0,1)$ the discount, and $\rho_0$ an initial-state distribution. $\mathcal{G}$ is a Dynamic Bayesian
Network (DBN)~\citep{murphy2002dynamic} over state factors $\rvs_t=(\rvs_t^1,\dots,\rvs_t^d)$, action $\rva_t=(\rva_t^1,\dots,\rva_t^m)$ and reward $r_t$. Let $\pa{X}$ denote the set of parents of node $X$ in $\mathcal{G}$. Hence, the state transitions are $\mathbb{P}_s(\rvs_t \mid \rvs_{t-1},\rva_{t-1})\;=\;\prod_{i=1}^d \mathbb{P}_s\!\big(\rvs_t^i \,\big|\, \pa{S_t^i}\big)$, where $\pa{S_t^i}\subseteq \{\rvs_{t-1}^1{:}\rvs_{t-1}^d,\,\rva_{t-1}^1{:}\rva_{t-1}^m\}$. The task-specific reward is a function of the parents of $R_t$: $\mathcal{R}(s_t,a_t)\;=\;\mathbb{E}[R_t \mid \pa{R_t}]$, where $\pa{R_t}\subseteq \{\rvs_t^1{:}\rvs_t^d,\,\rva_t^1{:}\rva_t^m\}$. The observation function is modeled as $\mathbb{P}_\rvo(\rvo_t \mid \rvs_t)\;=\;\prod_{j=1}^p \mathbb{P}_o\!\big(\rvo_t^j \,\big|\,\pa{\rvo_t^j}\big)$, where
$ \pa{\rvo_t^j}\subseteq \{\rvs_t^1{:}\rvs_t^d\}$.   

Then we consider a sequence of tasks $\{T_i\}_{i=1}^N$, where the latent state space is allowed to grow over time. Specifically, when the agent transitions to a new task, new state variables may be introduced that were absent in previous tasks.  Thus, in task $T_i$, the latent state is 
$\rvs_t^{i}=(\rvs_t^{i,1},\dots,\rvs_t^{i,d_i})$, where $d_1 \le \cdots \le d_N$, allowing the state dimensionality to
\emph{expand across tasks}. Each task is equipped with a task-specific DBN $\mathcal{G}^{(i)}$ over
$(\rvs_t^{i},\rva_t,\rvo_t, r^i_t)$. The corresponding state transition $\mathbb{P}_s^{(i)}$, observation function $\mathbb{P}_o^{(i)}$, and reward function $\mathcal{R}^{(i)}$ are also task-specific, yielding the task-specific DBNs~\citep{bilmes2000dynamic}.  Fig.~\ref{fig:example}(a) illustrates two tasks with different state–transition ($\rvs_t \rightarrow \rvs_{t+1}$) and state–reward ($\rvs_t \rightarrow r_{t}$). In addition, task 2 introduces an additional state factor $\rvs^{2,3}$.


\begin{Definition}[MIST indices for task $T_i$]
Let task $T_i$ have factored state
$\rvs^i_t=(\rvs_t^{i,1},\ldots,\rvs_t^{i,d_i})$
and a dynamic Bayesian network $\mathcal{G}^{(i)}$
with parent operator $\mathrm{pa}_{\mathcal{G}^{(i)}}(\cdot)$.
Define the current-time MIST indices
\begin{equation*}
I_{i,1}^{(t)} \coloneqq
\bigl\{ k \in [d_i] \mid
\rvs_t^{i,k} \in \mathrm{pa}_{\mathcal{G}^{(i)}}(r_t)
\bigr\},
\end{equation*}
and their one-step parent indices
\begin{equation*}
I_{i,2}^{(t)} \coloneqq
\bigl\{ k \in [d_i] \mid
\exists\, j \in I_{i,1}^{(t)} \text{ such that }
\rvs_{t-1}^{i,k} \in \mathrm{pa}_{\mathcal{G}^{(i)}}(\rvs_t^{i,j})
\bigr\}.
\end{equation*}
The MIST index set is
$U_i \coloneqq I_{i,1}^{(t)} \cup I_{i,2}^{(t)}$.
\end{Definition}

We provide formal definitions of sufficiency and minimality in Appendix~\ref{sec:defs}. 
An illustrative example of identifying the minimal sufficient states for each task is shown in Fig.~\ref{fig:example}(a). The MIST states for task 1 and 2 are $\{\rvs^{1,1}, \rvs^{1,2}\}$ and $\{\rvs^{2,2}, \rvs^{2,3}\}$, respectively, traced by red edges. 

\textbf{MIST for World Model \& Policy Learning.}
Given the MIST states, we can employ them directly for policy learning.  
For a concise comparison, Fig.~\ref{fig:example}(b) contrasts our approach with Dreamer-based methods~\citep{hafner2021mastering,hafner2025mastering}.  
While Dreamer learns unstructured latent states, we obtain compact, task-specific representations by projecting Dreamer-style latents through our structured decomposition and masking unrelated components.  
Consequently, our policy can be produced solely from the MIST states (pink dashed nodes in the right panel). 
Similarly, reward reconstruction uses only the reward’s parent variables (blue nodes), thereby respecting the identified structure.

\section{Method}
\begin{figure*}[h]
\begin{center}
\centerline{\includegraphics[width=.95\linewidth]{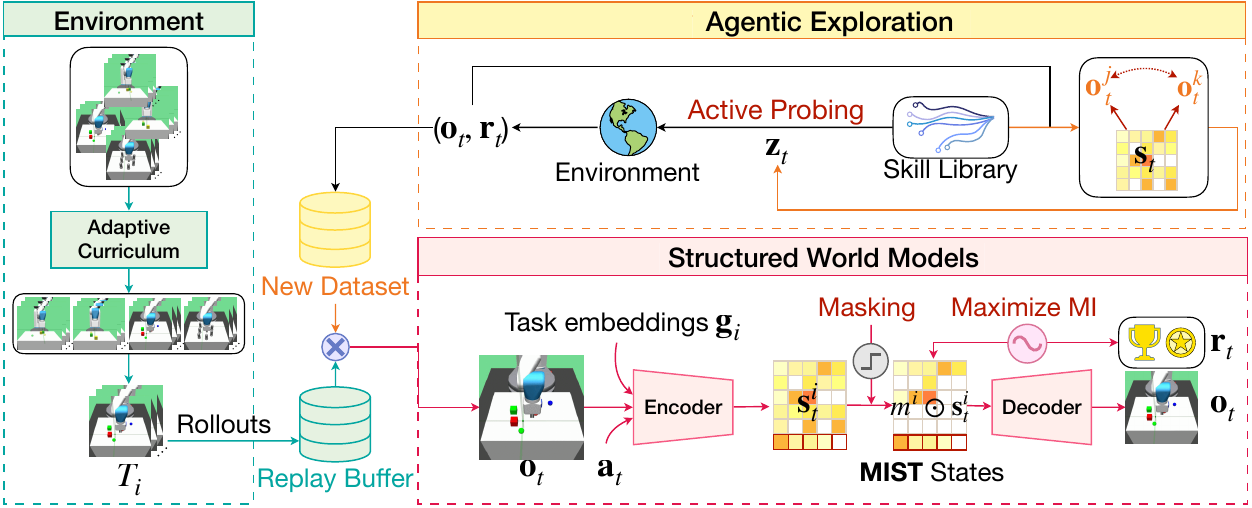}}
\caption{\textbf{Method overview.} 
On the agent side, for each new environment, the agent selects skills from its library to perform \textit{active probing}, collects informative data, under an adaptive curriculum. Conditioned on these informative data, the \textit{structured world model} is learned to distill the MIST states. 
}
\label{fig:pipeline}\vspace{-2mm}
\end{center}\vspace{-5mm}
\end{figure*}

With the formal definition of MIST states in place, we now present the algorithmic framework of learning \mist (Fig.~\ref{fig:pipeline}).
The framework consists of two \textit{interleaved} stages: \textit{agentic exploration} within the environment and \textit{structured world model learning}. In the remainder of this section, we first describe how the structured world model is learned. We then explain how, for each new task, agentic exploration discovers and collects informative interaction data for learning MIST representations. Finally, we introduce the adaptive curriculum, which facilitates agentic exploration by sequentially selecting tasks and their order.
\vspace{-2mm}
\subsection{Learning Structured World Models}\label{sec:worldmodel}\vspace{-2mm}
Our structured world model can be built on top of any off-the-shelf world-model backbone by learning a structured latent decomposition over its latent states. In this work, we instantiate our approach using Dreamer-V3~\citep{hafner2025mastering} as the base model.
For each task $T_i$, we learn a generative world model with the standard variational objective in Dreamer-V3~\citep{hafner2025mastering}:
\begin{equation}
\begin{aligned}
& \mathcal{L}^i_{\text{WM}}(\theta)
= \mathbb{E}_{\mathcal{D}}\Bigg[
 - \sum_{t=0}^{T-1}
   {\log p_\theta(\rvo_{t}\mid \rvs^i_{t})}  
 - {\log p_\theta(r^i_t, \gamma^i_t\mid \rvs^i_t)} \\ 
 &  +  {\vphantom{\sum}\beta\,\mathrm{KL}\!\big(q_\theta(\rvs^i_t\mid \rvo_{t},\rvs^i_{t-1}, \rva_{t})
 \,\|\, p_\theta(\rvs^i_t\mid \rvs^i_{t-1}, \rva_{t})\big)} 
\Bigg],
\end{aligned}\label{eq:dreamer}
\end{equation}
where $\theta$ denotes the world model parameters. However, this objective alone generally cannot identify the MIST states, since the learned latent space lacks explicit structure (left of Fig.~\ref{fig:example}(b)). We therefore incorporate a learnable mask $m^i \in (0,1)^d$ to select the task-specific subspace $\mathcal{S}^{\operatorname{min}}_i$ for task $T_i$: $m^i \odot \rvs^i_t \in \mathcal{S}^{\operatorname{min}}_i$, where $\rvs^i_t \in \mathbb{R}^d$ is the learned latent state from Dreamer.
In order to learn this mask, we have \textbf{Sufficiency Score}
$\mathcal{R}^{(i)}_{\text{suff}}(\theta)$ and the \textbf{Minimality Score}
$\mathcal{R}^{(i)}_{\mathrm{min}}(\theta)$ as follows.




The sufficiency score uses the masked states to maximize the likelihood of the observed rewards:
\begin{equation}
\mathcal{R}^{(i)}_{\text{suff}}
= \mathbb{E}_{\mathcal{D}}\!\left[
  \sum_{t=0}^{T-1}
  \log p_\theta\!\big(r_t \mid m^i \odot \rvs^i_t, \rva_t\big)
\right].
\label{eq:suffi}
\end{equation}
For the minimality score, we encourage the mask to retain only the necessary state dimensions by
maximizing the mutual information between the masked states and the task descriptor, while
penalizing the mask’s $\ell_1$ norm:
\begin{equation}
\mathcal{R}^{(i)}_{\mathrm{min}}
= \mathrm{MI}\!\big(m^i \odot \rvs^i_{1:T}, \rvg^i\big)
  - \lambda_M \,\|m^i\|_{1},
\label{eq:mini}
\end{equation}
where $\lambda_M$ is a balancing coefficient and $\rvg_i$ is the auxiliary task-specific information (e.g., task instructions). For the task descriptor,  when a simulator-provided task instruction is available, we encode this context using a pretrained RoBERTa~\citep{liu2019roberta} and use the resulting representation as the task embedding; otherwise, we use rewards only. Hence, in practice, it is either a pretrained task embedding or scalar reward information, depending on the benchmark. At test time, the same form of task information is provided as in training. Note that our method is a practical surrogate for the formal DBN-defined MIST states, aiming to recover a MIST-aligned task-specific subspace rather than exact latent DBN variables.

We use the world model with Dreamer-v3~\citep{hafner2025mastering} to obtain the latent
states, learn the soft mask $m^i$ via a gating-mask objective adapted from
\citet{rajamanoharan2024improving}, and estimate the mutual information term using
MINE~\citep{belghazi2018mutual}. Both approximated objectives are provided in Appendix~\ref{app:MI_mask}. {When moving to a new task (Fig.~\ref{fig:example}(a)), the dimensionality of the latent state is allowed to increase every time. Concretely, when transitioning from task $T_i$ to $T_{i+1}$, we fine-tune the encoder parameters learned for $T_i$ and adapt the final layer to accommodate the additional state dimensions required by $T_{i+1}$. This preserves the previously learned subspace while expanding the representation for new task-specific factors. Specific network details are in Appendix~\ref{app:grow}.}

With these structure-learning objectives, we can identify the MIST subspace from the learned
latents and restrict downstream control to this subspace; that is, both the policy and value function
operate on $\mathcal{S}^{\operatorname{min}}_i$. For task $T_i$, we optimize the cumulative return
$\mathcal{J}_{\mathrm{RL}}^{(i)}(\psi)
=
\mathbb{E}\Bigg[
  \sum_{t=0}^{T-1} \gamma^t\, r^i(\rvs^i_t,\rva_t)
\Bigg]$, with $\rva_t \sim \pi_\psi\big(\cdot \mid m^i \odot \rvs^i_t\big)$.

\vspace{-1mm}
\subsection{Agentic Exploration} 
\vspace{-1mm}
\label{sec:agent-exp}
Identifying MIST states requires informative interaction data for each task; without such data, the learned latents $\rvs^i$ cannot reliably capture the underlying task-relevant factors. Hence, we require informative trajectories that reveal how latent factors influence observations and rewards, rather than the offline datasets passively sampling a limited region of the state space. While on-policy exploration or other active-learning-curated data~\citep{pathak2017curiosity, sekar2020planning, kauvar2023curious} may occasionally provide such coverage, in novel environments, an agent without purposeful exploration will generally struggle to uncover task-relevant structure. Accordingly, upon entering a new environment, we enable such agentic exploration: the agent actively collects data through \textit{active probing}, executing probing skills $\rvz$ that induce controlled perturbations of the environment to differentiate latent factors.

The probing skills are designed to \textit{maximize} the \textit{information gain} about the world-model parameters $\theta$ in a new environment while ensuring broad, task-agnostic state coverage. They are modeled as high-level latent variables $\rvz$ sampled from a distribution $p_\eta$, with actions drawn from a skill-conditioned policy $\pi_\varepsilon(\rva \mid \rvs, \rvz)$. {Our objective can be decomposed into two consecutive procedures: (i) building a skill library, and (ii) selecting informative skills to probe the environment, enhancing information gain for world models.}

\textbf{Learning the Probing Skill Library $\mathcal{Z}$.}
Following the mutual information skill learning (MISL) framework~\citep{eysenbach2018diversity, park2024metra, zheng2025can}, we learn the \textit{exploration policy $\pi_\varepsilon(\rva\mid \rvs,\rvz)$} to maximize state–skill mutual information and action entropy. 

We select the prior distribution of skills as a uniform distribution over the d-dimensional unit hypersphere (a uniform von Mises–Fisher distribution~\citep{mardia2009directional},  $p(\rvz)=\operatorname{Unif}\left(\mathbb{S}^{d-1}\right)$) and learn the \textit{discriminator $q_\phi(\rvz\mid \rvs)$} to make skills identifiable from states.  We adopt off-the-shelf MISL algorithms; in particular, DIAYN~\citep{eysenbach2018diversity} and METRA~\citep{park2024metra}. Details are given in Appendix~\ref{app:misl}.

Specifically, we maximize the lower bound of mutual information with the objectives:
\begin{equation}
\max_{\varepsilon, \eta,\phi}\;
\mathbb{E}\Big[
\mathcal{H}\!\big(\pi_\varepsilon(\cdot\mid \rvs, \rvz)\big)
+
\log q_\phi(\rvz\mid \rvs)
-
\log p_\eta(\rvz)
\Big]\,
\label{eq:misl}
\end{equation}
where $\mathcal{H}$ computes the conditional entropy.
Empirically, we sample a skill $\rvz\!\sim\!p_\eta$ per $k$ steps, roll out $\pi_\varepsilon(\cdot\mid \rvs,\rvz)$, treat 
$\log q_\phi(\rvz\mid \rvs)$ as an intrinsic reward and maximize policy entropy analytically. To capture temporally extended effects, we replace $\rvs$ with a segment encoder $\phi_{\text{seg}}(\rvs_{t+1:t+k}, \rvs_t)$.


\textbf{Maximizing the Information Gain.}
Within the skill library, we select skills that maximize information gain, measured by their ability to facilitate the discovery of latent representations. Let $\rvs_t^{n_i}$
denote the $i$-th latent factor at time $t$, taking values in the space $\mathcal{V}_i$.
For two distinct values $v, v' \in \mathcal{V}_i$, under the
skill-conditioned exploration policy \(\pi_\varepsilon(\cdot \mid \rvs, \rvz)\), we define the $k$-step observation segments as
\begin{equation*}
\begin{aligned}
\rvo^{(v)}_{t:t+k}
&\sim \mathcal{P}\!\left(\,\rvo_{t:t+k}\;\middle|\; \rvs_t^{(n_i)}=v,\; \pi_\varepsilon(\cdot\mid \rvz)\right), \\
\rvo^{(v')}_{t:t+k}
&\sim \mathcal{P}\!\left(\,\rvo_{t:t+k}\;\middle|\; \rvs_t^{(n_i)}=v',\; \pi_\varepsilon(\cdot\mid \rvz)\right).
\end{aligned}
\end{equation*}
{These rollouts can be generated directly by the agent experimenting in the environment, e.g., lifting two cubes with different weights using different skills.}
A skill is informative if it induces observation segments whose distributions differ markedly for two distinct latent-factor values $v$ and $v'$. {When $\mathcal{V}_i$ is continuous, $v$ and $v'$ can be understood as the valuations of two independent samples from $\mathcal{V}_i$, chosen such that $v \neq v'$.} 
In particular, when $\rvo^{(v)}_{t:t+k}$ and $\rvo^{(v')}_{t:t+k}$ are clearly distinguishable, the resulting contrast reveals the latent factors responsible for the difference.  Hence, the information gain can be quantified as the separability of segments. 

We then encourage this separability of segments produced from different values of $\rvs_t^{(n_i)}$ via
a distance metric.
Specifically, we encode each sequence with $\phi_\xi(\cdot)$.
For each anchor \(\rvo^{v}_{t:t+k}\), we choose a positive sample \(\rvo^{(v^+)}_{t:t+k}\) from the same valuation of latent variables (i.e.,  $\rvs_t^{(n_i)}=v$ but with another rollout segment), and a set of negatives
\(\{\rvo^{(v')}_{t:t+k}\}_{v'\in\mathcal{N}(v)}\). We utilize an InfoNCE loss~\citep{oord2018representation}:
\begin{equation}
\begin{aligned}
\mathcal{L}_{\mathrm{cl}}^{(n_i)}(\varepsilon,\xi)
&= \mathbb{E}
\Bigl[
-\log
\frac{\operatorname{d}(\rvx_v, \rvx_{v^+})}
{\operatorname{d}(\rvx_v, \rvx_{v^+})
+ \sum_{v'\in\mathcal{N}(v)} \operatorname{d}(\rvx_v, \rvx_{v'})}
\Bigr].
\end{aligned}
\label{eq:contrastive}
\end{equation}
where $j^+$ denotes the positive index and $\mathcal{N}(j)$ the negative set. $\rvx$ is the encoded trajectory output under actions ${a_{t:t+T-1}\sim \pi_\varepsilon(\cdot \mid z)}$, and 
$\mathrm{d}(u,v) := \exp(\mathrm{cosine}(u,v))$.
Hence, the learned skills guide interaction to maximize expected information gain about task-relevant factors, yielding data most useful for world-model learning. Other than functioning as "scoring", this objective also updates the exploration policy $\pi_\varepsilon$ to make the agents grasp the skills that prioritize these behaviors.

\textbf{Adaptive Curriculum for Exploration.}
To facilitate the agents to collect informative data, we  use an adaptive curriculum that selects and orders environment–task pairs to expose the agent to informative transitions for world-model learning with progressively staging difficulty.

First, each environment provides minimal design scaffolding to ensure that such informative segments can be discovered.
For example, to encourage the discovery of object mass, we place two cubes with different weights on the workstation and allow the agent to infer the latent factors through interactions. Details are given in Appendix~\ref{app:env_design}. 

We then optimize the ordering of environment and tasks (a permutation $\sigma$ over $\{T_i\}_{i=1}^N$).
Since each task is instantiated in a specific environment, $\sigma$ is selected to maximize
\emph{data efficiency} for both world-model learning and policy learning. To operationalize it, we minimize world-model learning error and maximizing the cumulative rewards across all environments and tasks: 
\begin{equation*}
\min_{\sigma}\;
\sum_{j=1}^{N}\mathcal{L}_{\mathrm{WM}}\!\big(T_{\sigma(j)} \mid \mathcal{D}_{\sigma(<j)}\big)
\;-\;
\lambda_R \sum_{j=1}^{N}\mathcal{J}_{\mathrm{RL}}^{(\sigma(j))}(\psi).
\end{equation*}
Specifically, for world-model learning, since the latent dynamics are unknown,  following \cite{sekar2020planning}, we estimate the error via an ensemble
$\{f_\theta^{(m)}\}_{m=1}^{M}$ and define a disagreement. For downstream task learning, let $p_\theta(r\mid \rvs,\rva)$ be the reward head conditioned on the current learned latent states.
We use the reward loss to measure it directly. Hence, we have the evaluation proxy:
\vspace{-2mm}
\begin{equation}
\label{eq:disagreement}
\begin{aligned}
C_T(\mathcal{D})
&\;\coloneqq\;
\mathbb{E}_{(\rvs,\rva)\sim \mathcal{D}}
\Bigl[
d_{\mathrm{var}}\!\Big( \{f_\theta^{(m)}(\rvs,\rva)\}_{m=1}^{M} \Big) \\
&\hspace{6em}
-\log p_\theta(r \mid \rvs,\rva)
\Bigr].
\end{aligned}\vspace{-2mm}
\end{equation}
where $d_{\mathrm{var}}$ is the variance. 
This induces a natural notion of \emph{task difficulty}. 
Following Unsupervised Experiment Design (UED)~\citep{dennis2020emergent,rigter2024rewardfree}, we first explore all environments with the policy $\pi_\varepsilon$ and then have a buffer of per-task error scores $C_T(\mathcal{D})$. Each time, we select the next environment to reduce the worst-case error in learning. 

Concretely, we first explore all environments with the policy $\pi_\varepsilon$ and maintain a per-task error buffer $C_T(\mathcal{D})$. At each step, we choose the hardest environment, that is, the one with the largest current error, and collect additional data there:
$
T^\star \;=\; \arg\max_{T} \; C_T(\mathcal{D})$.
We then select $T^\star$ for learning. By repeating this procedure iteratively, we progressively reduce the maximum error across environments. In effect, this approximates solving a min–max objective, i.e., minimizing the maximum of the total error in the world models.

\vspace{-2mm}
\section{Related Works} \label{sec:rw}
\vspace{-1mm}
\textbf{Structures in World Model} Identifying meaningful structure in model-based reinforcement learning (MBRL) and world models is critical for generalization. Prior work can be broadly grouped into three lines. The first line focuses on compact state abstractions in (PO)MDPs. Representative examples include bisimulation-based methods~\citep{ferns2004metrics,castro2020scalable,zhang2021learning}, which aggregate states that are equivalent in return and transition dynamics under the abstraction. Other model-free approaches instead incorporate contrastive objectives to learn invariant features, as in CURL~\citep{laskin2020curl}. The second line aims to learn task- or control-aware representations in a model-based manner. Methods such as TIA~\citep{fu2021learning} and Denoised MDP~\citep{wang2022denoised} learn representations tailored to controllability or task-specific representations, thereby improving planning. The third line recovers structured world models by identifying graph structure in MDPs or POMDPs, i.e., learning factored MDPs~\citep{boutilier2000stochastic} from data. These approaches assume and exploit an underlying causal graph, either over observable states~\citep{pitis2020counterfactual,huang2021adarl, wang2022causal,feng2022factored, ji2024ace, chuck2025null, cao2025towards} or latent states~\citep{liu2023learning, hwang2024fine}, to support more generalizable decision making. Our work is most closely related to the last two lines but differs from them: rather than learning generic controllable representations or full graph structures, we explicitly target \emph{task-specific minimal sufficient} representations. In particular, we recover a factored structure but focus only on the components causally connected to the task, yielding compact, control-sufficient latents that can be integrated directly into policy learning.

\textbf{Unsupervised RL} aims to acquire meaningful behavior without external rewards, and prior work largely follows two directions. One direction focuses on \emph{intrinsic motivation}, where agents optimize surrogate signals that capture knowledge or competence about the environment. Examples include modeling prediction error as a novelty signal~\citep{burda2018exploration}, using curiosity-based measures~\citep{pathak2017curiosity,kauvar2023curious}, estimating disagreement among ensemble models~\citep{sekar2020planning}, and increasing empowerment or controllability across environments~\citep{eysenbach2018diversity,tiomkin2024intrinsic}. The other direction emphasizes \emph{unsupervised skill discovery}, where those methods maximize the mutual information between latent skills and states to learn diverse, meaningful behaviors~\citep{eysenbach2018diversity,park2024metra,hu2024disentangled,zheng2025can}. Our work is most closely aligned with the latter: the \emph{active intervention} component of \mist learns latent skills specifically to probe the environment and expose task-relevant latent factors, thereby enabling the identification of minimal, sufficient representations for downstream control. For curriculum learning, we draw inspiration from unsupervised environment design (UED)~\citep{dennis2020emergent,rigter2024rewardfree}, which provides curricula over sets of tasks for either policy learning or world-model learning. The work of \citet{dennis2020emergent} focuses on designing environments specifically for policy learning, while \citet{rigter2024rewardfree} studies reward-free cases. In contrast, our approach considers both settings simultaneously: we leverage the learned world model itself to progressively guide task selection to shape the curriculum.

\section{Experiments}
\begin{figure*}[t]
\begin{center}
\centerline{\includegraphics[width=.9\linewidth]{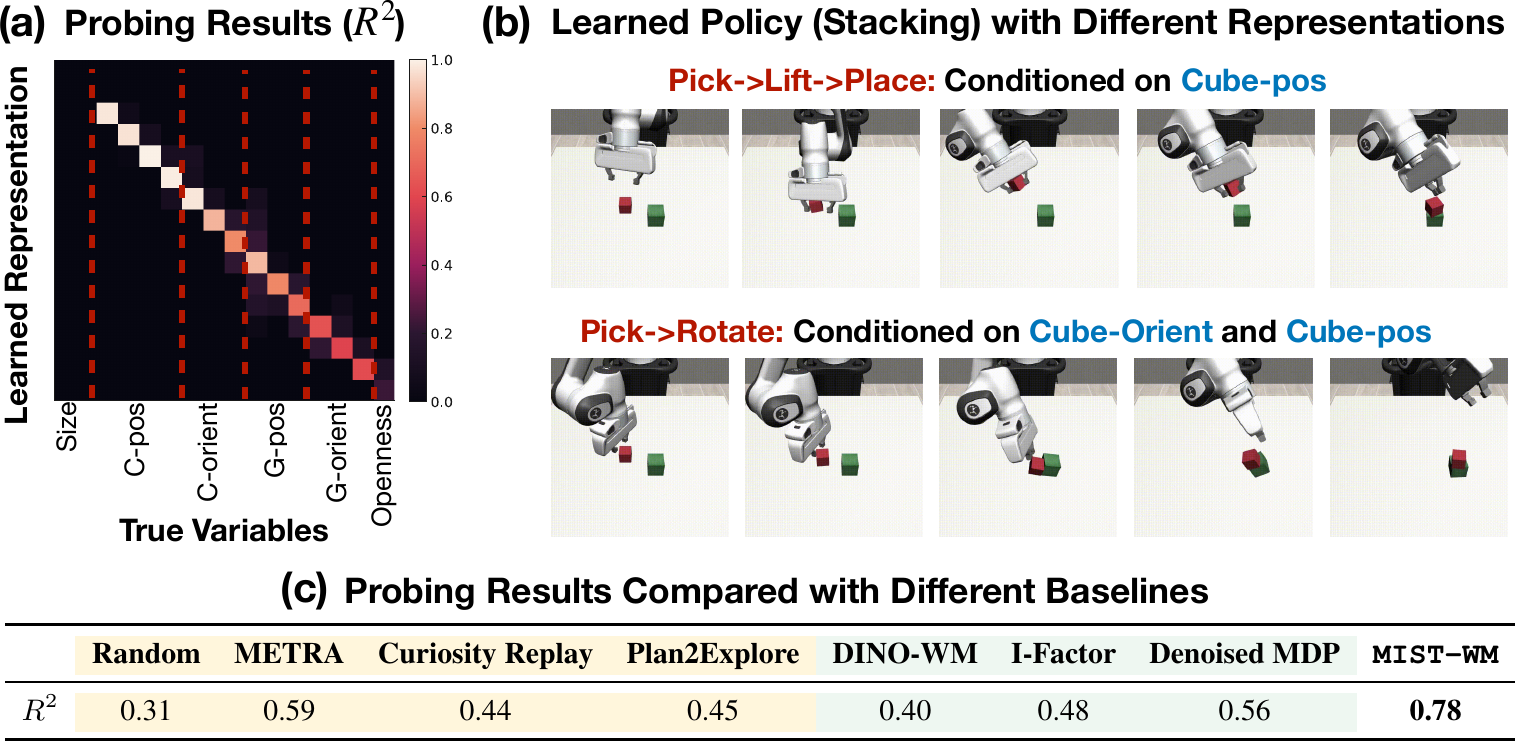}}
\caption{Results on Robosuite Stacking environment.  (a) The probing results ($R^2$) matrix ("C" and "G" mean "cube" and "gripper" respectively), and (b) Different policies conditioned on different learned representation variables. (c) Probing results on  Robosuite.}
\label{fig:res_1}
\end{center}
\end{figure*}
We evaluate \mist whether it can recover task-sufficient representations, and more importantly, whether these representations translate into improved downstream policy learning, particularly in terms of generalization.

\textbf{Benchmarks \& Baselines.} For a comprehensive evaluation, we test \mist on DMControl(Cheetah, Walker, Reacher in~\cite{tunyasuvunakool2020dm_control}), RoboSuite~\citep{zhu2020robosuite}, and Meta-World~\cite{yu2020meta}, and compare against baselines spanning the design space: world-model backbones (Dreamer-V3~\citep{hafner2025mastering}, DINO-WM~\citep{zhou2025dinowm}, TD-MPC2~\citep{hansen2023td}), factorization methods (Factored Dreamer~\citep{liu2023learning}), intrinsic-motivation approaches (Curiosity Replay~\citep{kauvar2023curious}, Plan2Explore~\citep{sekar2020planning}), and MISL methods (METRA~\citep{park2024metra}). 

\begin{beigebox}
\looseness=-5
{\textbf{RQ1: Representation Recoverability}}: \textit{Does the learned latent space recover minimal and task-sufficient representations that capture the essential factors for control? }
\end{beigebox}

We use a controlled RoboSuite setup (\cite{zhu2020robosuite}, Fig.~\ref{fig:res_1}a) {where ground-truth states are accessible in simulators, allowing us to directly assess whether the learned representation recovers the underlying latent factors.}
Specifically, we assess whether \mist can recover a minimal, task-sufficient subspace by mapping learned coordinates to ground truth via linear probes and reporting alignment $R^2$ (coefficient of determination \cite{wright1921correlation}).{The procedure for mapping the learned representations to ground-truth variables is detailed in Appendix~\ref{app:rsquare}.}
We also compare the learned representations with ablations and baselines through analyses along two axes: 
\emph{(i) Data collection:} Random (matched budget), MISL-only (METRA~\citep{park2024metra}), and intrinsic-reward methods (Curiosity Replay~\citep{kauvar2023curious} and Plan2Explore~\citep{sekar2020planning}).
\emph{(ii) World model:} using foundation-model (DINO-WM~\citep{zhou2025dinowm}) and factorized models (I-Factor~\citep{liu2023learning}, Denoised MDP~\citep{wang2022denoised}).
\begin{figure*}[h]
\begin{center}
\centerline{\includegraphics[width=\linewidth]{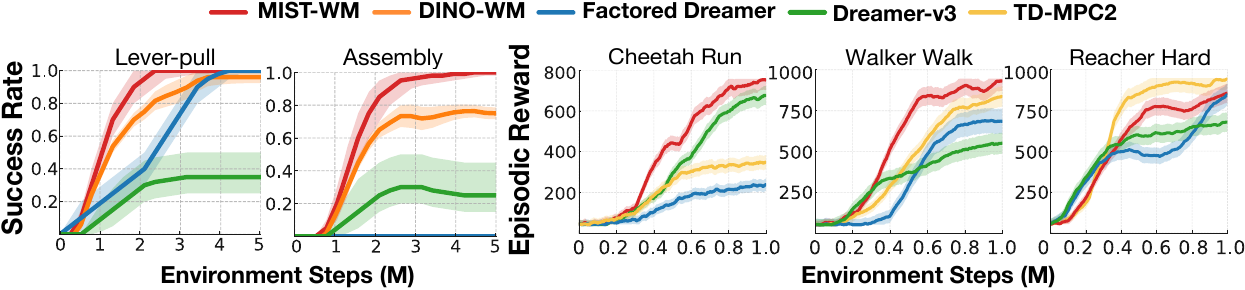}}
\caption{Results (across 5 seeds) on single-task learning, including \textbf{robotic manipulation} tasks (Meta World) and \textbf{locomotion} (DMControl). Shaded areas indicate the standard errors.}
\label{fig:res_single_task_short}
\end{center}
\end{figure*}

{\textbf{Take-Away 1: \mist can recover task-relevant representations.}} As shown in Fig.~\ref{fig:res_1}(a), we obtain near one-to-one alignment for task-relevant factors, including cube position/orientation and gripper position/orientation. However, the static or non-intervenable variables (e.g., object size) are not identified. Likewise, very fine-grained cues (i.e., the gripper open/close states) are not identified: though such signals exist in the data, they are hard to capture with contrastive pairs. In general, \mist identifies a compact subspace that retains what matters for the task and discards others. 
Fig.~\ref{fig:res_1}(c) shows the average $R^2$ (diag) and Appendix Table~\ref{tab:representation} gives the full results and Appendix~\ref{app:rsquare} gives the calculation details.
We find that learned \mist attains the highest alignment. Within different data curation methods, aside from ours, METRA yields the best alignment, while intrinsic-reward methods are comparable.
For world models, structural representation learning (I-Factor and Denoised MDP) outperforms DINO-WM on average, suggesting the effectiveness of learning disentangled representations of these two.

{\textbf{Take-Away 2: Different learned state representations induce qualitatively different policy behaviors}}. In RoboSuite Stacking (Fig.~\ref{fig:res_1}(c)), conditioning the policy on position features alone yields an optimal pick–lift–place sequence. In contrast, conditioning on both orientation and position enables a shortcut: when the objects are close, the agent can directly rotate one object onto the other. This indicates that an aligned latent space provides more controllable and interpretable behaviors.  

\begin{figure*}[t]
\begin{center}
\centerline{\includegraphics[width=\linewidth]{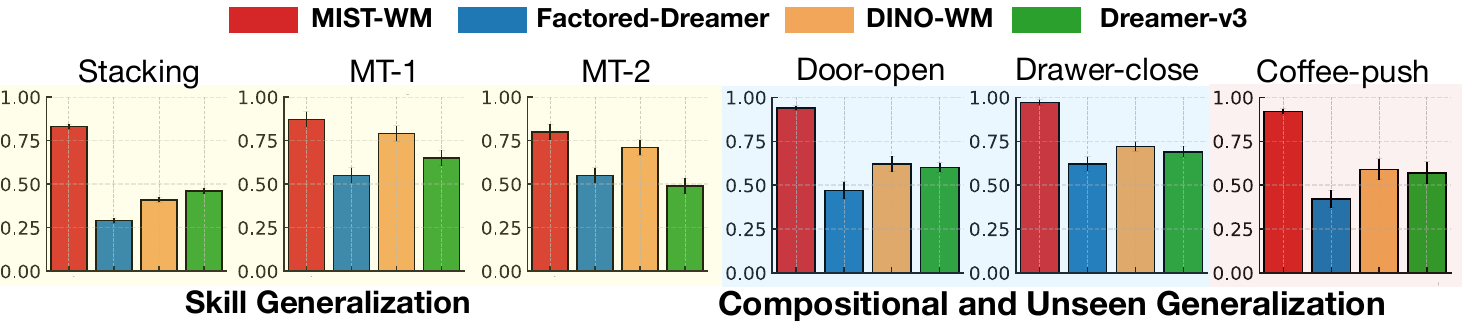}}
\caption{Success rate (across 5 seeds) on skill and compositional generalization, including tasks in Stacking (Robosuite), MT-1 \& 2 (kitchen), and Door-open, Door-close, Coffee-push (Meta-World). {Error bars indicate the standard errors.}}
\label{fig:res_gen_short}
\end{center}
\end{figure*}
\begin{beigebox}
\looseness=-5
{\textbf{RQ2: Policy Learning}}: \textit{Given a fixed task sequence, do policies trained on MIST states achieve strong sample efficiency relative to baseline world models?}
\end{beigebox}

We consider evaluating the single task policy learning  (\textit{no curriculum, only with MIST states learning and active probing skills}) on benchmarks, including robotic control (Meta-World~\citep{yu2020meta}) and locomotion tasks from DMControl~\citep{tunyasuvunakool2020dm_control}. 
We compare against DINO-WM~\citep{zhou2024dino}, I-Factor (a Dreamer-based factored model; labeled \emph{Factored Dreamer})~\citep{liu2023learning}, Dreamer-V3~\citep{hafner2025mastering}, and TD-MPC2~\citep{hansen2024tdmpc}. Regarding policy learning, for DMControl and Franka-kitchen, all methods use Proximal Policy Optimization (PPO, \cite{schulman2017proximal}), whereas for Meta-World and RoboSuite, they use Soft Actor-Critic (SAC, ~\cite{haarnoja2018soft}). 
Results are in Fig.~\ref{fig:res_single_task_short} with full results  in Appendix  Fig.~\ref{fig:res_single_task}, Table \ref{tab:mt_2m}-\ref{tab:dmc_1m}. Computational cost analysis is given in Appendix \ref{app:cost}.
\begin{figure}[t]
\begin{center}
\centerline{\includegraphics[width=\linewidth]{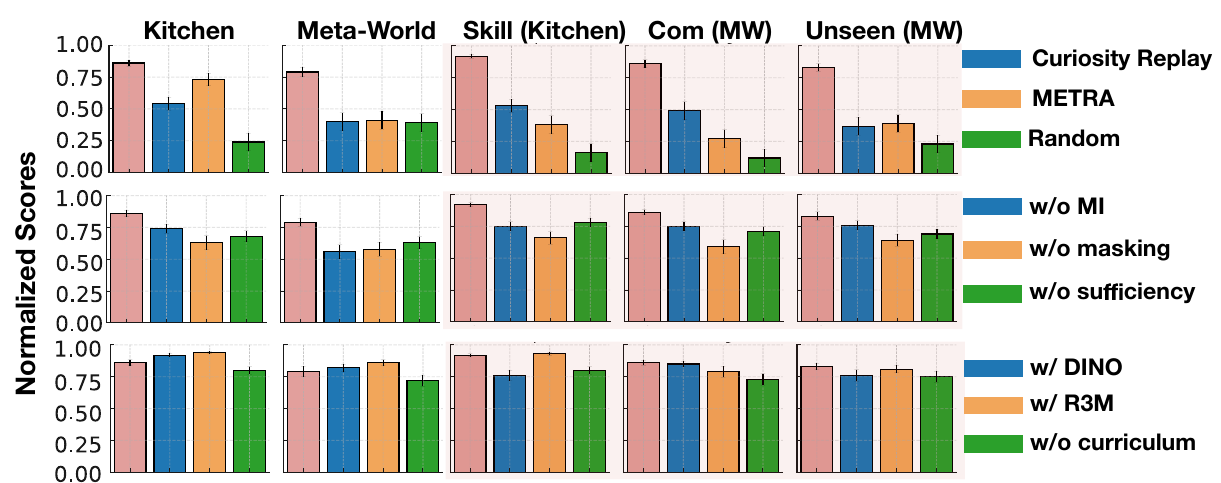}}
\caption{\textbf{Ablation studies}. We consider studies on exploration policy (Row 1); World model terms (Row 2); and Backbones and curriculum (Row 3). {Pink bars are original \mist and error bars indicate the standard errors.}}
\label{fig:res_ablation_short}
\end{center}\vspace{-1cm}
\end{figure}

{\textbf{Take-Away 3: \mist consistently improves data efficiency and policy learning performances across most tasks, except for Reacher-Hard.}}
On Meta-World, DINO-WM is typically the second-best, while Factored Dreamer and Dreamer-V3 often lag behind DINO-WM. 
On locomotion, \mist outperforms TD-MPC2 on Cheetah and Walker but underperforms on Reacher-Hard, while still exceeding Dreamer-based baselines. 
We attribute this to differing objectives: TD-MPC optimize a model-free control objective, whereas \mist builds on Dreamer objectives.
\begin{beigebox}
\looseness=-5
{\textbf{RQ3: Sample-Efficiency and Generalization}}:\textit{ Do the learned task-sufficient representations transfer beyond the training tasks, enabling systematic generalization in policy learning?}
\end{beigebox}

We evaluate \mist on three types of generalization using the learned MIST representations. Summary results are in Fig.~\ref{fig:res_gen_short} and full results are in Fig.~\ref{fig:res_gen}. 

{\textbf{Take-Away 4: The learned MIST representations consistently improve skill, compositional, and unseen-task generalization.}}
\emph{(i) Skill generalization.}
{Can the learned MIST states facilitate better skill-level generalization?}
We study this in RoboSuite~\citep{zhu2020robosuite} and Franka-Kitchen~\citep{gupta2019relay}, where source tasks are learned and then transferred to target tasks that require composing multiple the learned representations. 
Results show that \mist achieves large improvements on both source and target tasks, with especially pronounced gains in RoboSuite target tasks and harder Kitchen tasks (e.g., Kitchen-light). 
This suggests that the learned representation help  scale to complex, long-horizon behaviors.  
\emph{(ii) Compositional generalization.}  
Here we combine learned “objects’’ and “skills’’ to create novel tasks. 
In Meta-World~\citep{yu2020meta}, we train on (door-unlock, drawer-open, faucet-close, handle-pull) and evaluate on (door-open, drawer-close). 
\mist outperforms all baselines on both base and target tasks, with especially strong gains on the door-open target.  \emph{(iii) Unseen-task generalization.}
Finally, in Meta-World, we test on coffee-push, a task where neither “coffee’’ nor “push’’ appeared during training. \mist generalizes well, indicating that the learned latent representations capture transferable structure that extends to previously unseen factors. {We learn an expandable state space that supports new tasks. When a task involves new combinations of skills or objects, the agent can initialize from previously learned states and learn the corresponding policy with fewer samples.}

\textbf{> Ablation Studies} To evaluate each component of \mist, we conduct ablations along three axes:
\emph{(i) Different policy} for data curation: \emph{Random}, \emph{Intrinsic motivation} (Curiosity Replay~\citep{kauvar2023curious}), and \emph{MISL} (METRA);
\emph{(ii) World Model terms}: removing the MI term in the \emph{minimality} score, removing \emph{masking} in the minimality score, and removing the \emph{sufficiency} term; and
\emph{(iii) Backbone and curriculum}: {replacing the Dreamer-v3 encoder with \emph{DINO-WM} or \emph{R3M}}, and training w.r.t. a \emph{no-curriculum} variant (i.e., random order).
We report both single-task learning and cross-environment generalization (Fig.~\ref{fig:res_ablation_short}, full results are in Appendix Fig.~\ref{fig:res_ablation}). All scores are normalized to an \emph{Oracle} that directly receives ground-truth states in each environment.

{\textbf{Take-Aways for Ablations:}} The default \mist achieves the best overall.
In generalization settings (pink panels), performance approaches the Oracle (often $\approx 0.9$), and exceeds single-task learning, indicating that the \textit{learned representations transfer across tasks}.  
For case (i), METRA is typically second-best, while Random is consistently worst.
On generalization tasks, Curiosity Replay sometimes surpasses METRA, consistent with its broader, task-agnostic coverage, improving transfer. For case (ii), 
Removing MI, masking, or sufficiency uniformly degrades performance.
The effects are similar in single-task settings; in generalization, masking has the largest impact, suggesting that enforcing compactness improves transfer. For case (iii), using DINO-WM~\citep{zhou2024dino} or R3M~\citep{nair2023r3m} features can help on some tasks, but is not consistently superior.
Removing the curriculum causes larger drops in generalization than in single-task learning, reflecting the curriculum’s role in generalization across tasks. Results indicate that synergy between agentic exploration (probing skills) and structured models is crucial, as the absence of either component leads to degraded information acquisition and suboptimal learning dynamics.

\vspace{-2mm}
\section{Conclusion}
\vspace{-1mm}
We study how agents can learn world models whose latent spaces provide task-sufficient representations. Our goal is to distill compact latent representations directly from pixels, focusing the agent on task-relevant factors. To this end, we synergize agentic exploration with structured world-model learning in a closed-loop manner: agentic exploration curates informative interaction data through active probing under an adaptive curriculum, while structure-aware learning objectives leverage these data to distill minimal and sufficient representations in the world model, which in turn guide subsequent exploration.
This provides a principled way to learning meaningful representations for world models, enabling efficient and generalizable policy learning.




\section*{Impact Statement}

This paper presents work whose goal is to advance the field of Machine
Learning. There are many potential societal consequences of our work, none of 
which we feel must be specifically highlighted here.

\section*{Acknowledgment}

We would like to acknowledge the support from NSF Award No.~2229881, AI Institute for Societal Decision Making (AI-SDM), the National Institutes of Health (NIH) under Contract R01HL159805, and grants from Quris AI, Florin Court Capital, MBZUAI-WIS Joint Program, and the Al Deira Causal Education project.

\bibliography{ref.bib}
\bibliographystyle{iclr2026_conference}
\clearpage
\onecolumn

\begin{center}
\Large{{\textbf{Appendix}}}
\end{center}
\vspace{4mm}
\startcontents[sections]
\printcontents[sections]{l}{1}{\setcounter{tocdepth}{3}}
\startcontents
    \setcounter{figure}{0}
    \renewcommand{\thefigure}{A\arabic{figure}}
    \setcounter{table}{0}
    \renewcommand{\thetable}{A\arabic{table}}
    \renewcommand{\thesection}{\Alph{section}}
    \renewcommand{\theequation}{A\arabic{equation}}
    \setcounter{equation}{0}
\appendix
\section{Limitations and Future Works}
Several limitations remain. Most notably, our experiments are confined to simulated domains; extending the framework to real-world systems, such as physical robotic platforms, is an important direction for future work. Although our methodology is model-agnostic, scaling it to more expressive and scalable world-model architectures, such as diffusion-based world models~\citep{wan2025, chi2025wow, guo2025ctrlworldcontrollablegenerativeworld} will be left for future works.
\section{Notations and Concepts}
\subsection{Definitions of Minimality and Sufficiency}\label{sec:defs}


\begin{Definition}[Sufficiency]
The index set $U_i$ is \emph{sufficient for} $T_i$ if
\begin{equation}
r^i_t \perp\!\!\!\perp
(\rvs^i_{t-1},\rvs^i_t)\setminus
\bigl(\rvs^i_{t-1}|_{I_{i,2}^{(t)}},\rvs^i_t|_{I_{i,1}^{(t)}}\bigr)
\mid
\bigl(\rva_t,\rvs^i_{t-1}|_{I_{i,2}^{(t)}},\rvs^i_t|_{I_{i,1}^{(t)}}\bigr).
\end{equation}
Equivalently, given $\rva_t$, the reward depends on the state only through the current MIST states and their one-step parents.
\label{def: suff}
\end{Definition}


\begin{Definition}[Minimality]
A sufficient index set $U_i$ is \emph{minimal} if there exist no strict subsets
$J^{(t)} \subsetneq I_{i,1}^{(t)}$ and
$J^{(t-1)} \subsetneq I_{i,2}^{(t)}$
such that
\begin{equation}
r^i_t \perp\!\!\!\perp
(\rvs^i_{t-1}, \rvs^i_t)\setminus
\bigl(\rvs^i_{t-1}|_{J^{(t-1)}}, \rvs^i_t|_{J^{(t)}}\bigr)
\mid
\bigl(\rva_t, \rvs^i_{t-1}|_{J^{(t-1)}}, \rvs^i_t|_{J^{(t)}}\bigr).
\end{equation}
Equivalently, removing any variable from $U_i$ destroys sufficiency.
\label{def:mini}
\end{Definition}
\subsection{Notations} 
Table~\ref{tbl:notations} collects the notations used in the theorem proofs for clarity and consistency. 

\begin{table}[htp!]
\caption{List of notations, explanations, and corresponding value ranges.}
\centering
\resizebox{\textwidth}{!}{
\begin{tabular}{cll}
\toprule
$\textbf{Variable}$ & $\textbf{Explanation}$ & \textbf{Support} \\
\midrule
$\mathcal{S}$ & (latent) state space & $\mathcal{S} \subseteq \mathbb{R}^{d_s}$ \\
$\mathcal{A}$ & action space & $\mathcal{A} \subseteq \mathbb{R}^{d_a}$ \\
$\mathcal{O}$ & observation space & $\mathcal{O} \subseteq \mathbb{R}^{d_o}$ \\
$\mathcal{D}$ & dataset / replay buffer distribution & trajectories over $(\rvs,\rva,\rvo,r)$ \\
$\mathcal{Z}$ & skill library (set of learnable skills) & $\mathcal{Z} \subseteq \mathbb{R}^{d_z}$ \\
$\gamma$ & discount factor & $\gamma \in (0,1)$ \\
$\rho_0$ & initial-state distribution & $\rho_0: \mathcal{S} \to [0,1]$ \\
$\{T_i\}_{i=1}^N$ & set of tasks & $i=1,\dots,N$ \\
$\rvs_t^i=(\rvs_t^{i,1},\ldots,\rvs_t^{i,d_i})$ & factored (learned) state of task $i$ & $d_i$ state factors \\
$I_i^{(t)}$ & indices of state factors influencing reward $r_t$ & $I_i^{(t)} \subseteq [d_i]$ \\
$I_i^{(t-1)}$ & indices of one-step parents of $I_i^{(t)}$ & $I_i^{(t-1)} \subseteq [d_i]$ \\
$U_i$ & union of current and one-step parents & $U_i = I_i^{(t)} \cup I_i^{(t-1)}$ \\
$\rvs^i_t|_{I_i^{(t)}}$ & current-time MIST states & subset of $\rvs^i_t$ \\
$\rvs^i_{t-1}|_{I_i^{(t-1)}}$ & one-step parent MIST states & subset of $\rvs^i_{t-1}$ \\
$\rvg_i$ & task descriptor / embedding for $T_i$ (can be rewards in simple cases) & $\rvg_i \in \mathbb{R}^{d_g}$ \\
$m^i$ & soft/binary mask selecting task-specific subspace for $T_i$ & $m^i \in (0,1)^{d_i}$ \\
$\mathcal{S}^{\operatorname{min}}$ & minimal sufficient subspace for task $T_i$ & $\mathcal{S}^{\operatorname{min}} \subseteq \mathcal{S}$ \\
$T$ & horizon length & $T \in \mathbb{N}^+$ \\

$\rvz$ & latent skill variable & $\rvz \in \mathbb{R}^{d_z}$ \\

$\mathcal{N}(v)$ & negative sample index set for contrastive learning & $\mathcal{N}(v) \subseteq \{1,\dots,N\}$ \\
$\rvx_v, \rvx_{v^+}, \rvx_{v'}$ & encoded representations of anchor, positive, and negative samples & $\rvx \in \mathbb{R}^{d_h}$ \\
\midrule
$\textbf{Function}$ & & \\
\midrule
$\mathcal{G}$ & dynamic Bayesian network (DBN) over $(\rvs_t,\rva_t,r_t)$ & graph structure \\
$\mathcal{G}^{(i)}$ & task-specific DBN over $(\rvs_t^{i}, \rva_t, \rvo_t, r_t^i)$ & graph structure \\
$\mathbb{P}_s(\rvs_t \mid \rvs_{t-1}, \rva_{t-1})$ & state transition probability & $\mathcal{S} \times \mathcal{A} \to \Delta(\mathcal{S})$ \\
$\mathbb{P}_o(\rvo_t \mid \rvs_t)$ & observation function & $\mathcal{S} \to \Delta(\mathcal{O})$ \\
$\mathcal{R}(\rvs_t,\rva_t)$ & reward function & $\mathcal{S}\times\mathcal{A} \to \mathbb{R}$ \\
$\mathbb{P}_s^{(i)}(\rvs_t \mid \rvs_{t-1}, \rva_{t-1})$ & transition probability of task $i$ & $\mathcal{S} \times \mathcal{A} \to \Delta(\mathcal{S})$ \\
$\mathbb{P}_o^{(i)}(\rvo_t \mid \rvs_t)$ & observation function of task $i$ & $\mathcal{S} \to \Delta(\mathcal{O})$ \\
$\mathcal{R}^{(i)}(\rvs_t,\rva_t)$ & reward function of task $i$ & $\mathcal{S}\times\mathcal{A} \to \mathbb{R}$ \\
$\mathrm{pa}_{\mathcal{G}^{(i)}}(\cdot)$ & parent operator in task-specific DBN $\mathcal{G}^{(i)}$ & node $\mapsto$ set of parents \\
$\mathcal{L}^i_{\text{WM}}(\theta)$ & task-conditioned world model objective for $T_i$ & variational training loss \\
$\mathcal{R}^{(i)}_{\text{suff}}(\theta)$ & sufficiency score  & ${\mathbb{E}_{\mathcal{D}}\!\left[ \sum_{t=0}^{T-1} \log p_\theta\!\big(r_{t} \mid m^i \odot \rvs_{t}^{i}, \rva_{t}\big)\right]}$ \\
$\mathcal{R}^{(i)}_{\mathrm{min}}(\theta)$ & minimality score for selecting $\mathcal{S}^{\operatorname{min}}$ & model-dependent criterion \\
$p_\theta\!\big(r_{t} \mid m^i \odot \rvs_{t}^{i}, \rva_{t}\big)$ & conditional reward likelihood (task $i$) & scalar density \\
$\mathcal{J}_{\mathrm{RL}}^{(i)}(\psi)$ & expected cumulative return objective for task $i$ & $\mathbb{E}\!\left[\sum_{t=0}^{T-1} \gamma^t r^i(\rvs_t^i,\rva_t)\right]$ \\
$\pi_\psi(\cdot \mid m^i \odot \rvs_t^{i})$ & task-conditioned policy based on masked latent states & distribution over $\mathcal{A}$ \\
$p_\eta$ & skill prior distribution & $\rvz \sim p_\eta$ \\
$\pi_\varepsilon(\rva \mid \rvs,\rvz)$ & skill-conditioned policy & distribution over $\mathcal{A}$ \\
$p(\rvz)$ & uniform skill prior on unit hypersphere & $\operatorname{Unif}(\mathbb{S}^{d-1})$ \\
$q_\phi(\rvz \mid \rvs)$ & discriminator for skill identification from states & $\mathcal{S} \to \Delta(\mathcal{Z})$ \\
$\phi_{\text{seg}}(\rvs_{t+1:t+k}, \rvs_t)$ & segment encoder mapping state segments to embeddings & $\mathbb{R}^{(k+1)\times d_s} \to \mathbb{R}^{d_h}$ \\
$\phi_\xi(\cdot)$ & sequence encoder for contrastive learning & trajectories $\to \mathbb{R}^{d_h}$ \\
$\mathcal{L}_{\mathrm{contrast}}^{(n_i)}$ & InfoNCE-style contrastive loss for skill discrimination & scalar objective \\
$\operatorname{d}(\rvx,\rvx')$ & similarity/distance function between embeddings & $\mathbb{R}^{d_h}\times\mathbb{R}^{d_h}\to \mathbb{R}$ \\
$C_T(\mathcal{D})$ & evaluation proxy combining disagreement and reward likelihood & $\mathbb{E}_{(\rvs,\rva)\sim \mathcal{D}}[\cdot]$ \\
$d_{\mathrm{var}}(\{f_\theta^{(m)}\}_{m=1}^M)$ & variance-based disagreement measure across ensemble predictions & $\mathbb{R}^M \to \mathbb{R}$ \\
$f_\theta^{(m)}(\rvs,\rva)$ & $m$-th ensemble dynamics model prediction & $\mathcal{S}\times\mathcal{A}\to\mathbb{R}$ \\
\midrule
$\textbf{Symbol}$ & & \\
\midrule
$\odot$ & Hadamard (element-wise) product & $(\mathbb{R}^{d}\times\mathbb{R}^{d}) \to \mathbb{R}^{d}$ \\
$\mathbb{S}^{d-1}$ & $(d-1)$-dimensional unit sphere & $\{ \rvz \in \mathbb{R}^d : \|\rvz\|_2 = 1\}$ \\
\bottomrule
\end{tabular}
}
\label{tbl:notations}
\end{table}

\subsection{{Concepts}}
Here we explain a few key concepts used in our framework as a complement of the main paper. 
\paragraph{Unstructured latent states}
We instantiate the world model with a recurrent state-space model (RSSM) as in
Dreamer-v3, which produces latent states
$\rvs_t \in \mathbb{R}^d$ from observations $\rvo_t$ and actions $\rva_t$ via
\[
q_\theta(\rvs_{t+1} \mid \rvo_{t}, \rvs_{t}, \rva_{t}), \qquad
p_\theta(\rvo_t \mid \rvs_t), \quad
p_\theta(\rvs_{t+1} \mid \rvs_t, \rva_t).
\]
These latent states are \emph{unstructured} in the sense that no explicit factorization or
task-specific subspace is imposed.

\paragraph{Structure learning}
On top of the unstructured latents, we perform \emph{structure learning} to extract a
task-specific subspace. For each task $T_i$, let $\rvs^i_t \in \mathbb{R}^d$ denote the
corresponding Dreamer latent and $\rvg^i$ a task descriptor. We learn a mask
$m^i \in (0,1)^d$ and a factored generative model over $(\rvs^i_t,\rvo_t,r^i_t)$ conditioned on
$\rvg^i$ to optimize Eq.~\ref{eq:suffi}-\ref{eq:mini} to encourage a low-dimensional, task-sufficient subspace. 

\paragraph{MIST states}
For task $T_i$, the \emph{Minimal, Sufficient, Task-specific (MIST) states} are the masked
latents
\[
\tilde{\rvs}^i_t \;=\; m^i \odot \rvs^i_t \;\in\; \mathcal{S}^{\mathrm{min}}_i,
\]
where $\mathcal{S}^{\mathrm{min}}_i \subseteq \mathbb{R}^d$ denotes the minimal sufficient
subspace recovered by structure learning. Intuitively, MIST states are those coordinates of
$\rvs^i_t$ that are (i) sufficient for predicting rewards $r^i_t$ given $\rva_t$, and (ii) cannot
be removed without violating sufficiency.

Concretely, the connection between \emph{masked states}, \emph{latent factors}, and the \emph{expandable state space} is as follows. Dreamer provides an unstructured latent vector $\rvs^i_t \in \mathbb{R}^{d_i}$ for each task $T_i$, whose coordinates we interpret as latent factors. As we move to new tasks and the environment introduces additional mechanisms, we allow the latent dimensionality $d_i$ to grow (expandable state space). For each task $T_i$, we then learn a soft mask $m^i \in (0,1)^{d_i}$ that gates these coordinates and defines the task-specific subspace $m^i \odot \rvs^i_t$, which we call the MIST states. Thus:
(i) \emph{latent factors} are the coordinates of $\rvs^i_t$,
(ii) the \emph{expandable state space} means that the dimensionality $d_i$ can increase across tasks, and
(iii) the \emph{masked states} $m^i \odot \rvs^i_t$ are the subset of those factors that our structure-learning objectives identify as minimal and sufficient for predicting rewards.

\paragraph{Active probing skills}
To obtain informative data for structure learning, the agent executes \emph{probing skills}
parameterized by a latent variable $\rvz$. We maintain a skill prior $p_\eta(\rvz)$ (e.g.,
uniform on a hypersphere) and a skill-conditioned exploration policy
\[
\rva_t \sim \pi_\varepsilon(\cdot \mid \rvs_t, \rvz),
\]
trained to maximize a mutual-information style objective (Eq.~\ref{eq:misl}) and a contrastive objective (Eq.~\ref{eq:contrastive}). Rolling
out $\pi_\varepsilon$ yields \emph{active probing} that probe how latent factors affect
observations and rewards, producing data that is informative for discovering MIST states.

\paragraph{Segment separability}
Given a latent factor $s^{(n)}_t$ taking values in some space $\mathcal{V}_n$, and a fixed skill
$\rvz$, we consider $k$-step observation segments
\[
\mathbf{o}^{(v)}_{t:t+k} \sim P\big(\rvo_{t:t+k} \mid s^{(n)}_t = v,\; \pi_\varepsilon(\cdot \mid \cdot,\rvz)\big),
\qquad
\mathbf{o}^{(v')}_{t:t+k} \sim P\big(\rvo_{t:t+k} \mid s^{(n)}_t = v',\; \pi_\varepsilon(\cdot \mid \cdot,\rvz)\big),
\]
for two distinct values $v \neq v' \in \mathcal{V}_n$. We encode segments with
$\mathbf{x}^{(v)} = \phi_\xi(\mathbf{o}^{(v)}_{t:t+k})$ and define \emph{segment separability} via an
InfoNCE-style contrastive loss.
Skills that induce highly separable segments (low $\mathcal{L}^{(n)}_{\mathrm{contrast}}$) provide
strong evidence about the underlying latent factors and are therefore preferred for active probings.

\section{Implementation Details on \mist}
\subsection{Mutual Information and Masking Learning}\label{app:MI_mask}
We use MINE~\citep{belghazi2018mutual} to estimate the mutual information.
Following \cite{wang2025modeling}, to avoid instability from inaccurate estimates early in training, the MI weight is annealed from a small value to its target with a cosine schedule:
\begin{equation}
\label{eq:mi-schedule}
\alpha_{\text{MI}}(t)
=
\begin{cases}
\alpha_{\text{start}} + \dfrac{1}{2}\!\left(\alpha_{\text{end}}-\alpha_{\text{start}}\right)\!\left(1-\cos(\pi t)\right), & \alpha_{\text{end}}>\alpha_{\text{start}},\\[6pt]
\alpha_{\text{end}} + \dfrac{1}{2}\!\left(\alpha_{\text{start}}-\alpha_{\text{end}}\right)\!\left(1+\cos(\pi t)\right), & \text{otherwise},
\end{cases}
\end{equation}
where \(t\in[0,1]\) is normalized training time. This smoothly activates the MI constraint and mitigates variance-driven min–max oscillations.

To induce compact latents without the shrinkage artifacts of plain \(\ell_1\) penalties, we use a modified gated mask inspired by \citet{wang2025modeling, rajamanoharan2024improving}. 
The mask is applied to the latent state $\rvs$; the stochastic component is left unchanged.
Let $\rvs$ denote the latent vector. We define a binary gate and a magnitude reparameterization:
\begin{align}
\text{Gate:}\quad 
\mathrm{Gate}(\rvs) &\;\coloneqq\; \mathbf{1}\!\left(\,|\rvs| + b_{\text{gate}} > 0\,\right),
\label{eq:gate}
\\[4pt]
\text{Magnitude:}\quad 
f_{\operatorname{mag}}(\rvs) &\;\coloneqq\; \exp(r_{\text{mag}})\,\rvs + b_{\text{mag}},
\label{eq:mag}
\end{align}
and the masked latent is
\begin{equation}
\label{eq:masked}
\vm \odot \rvs \;\coloneqq\; \mathrm{Gate}(\rvs)\; \odot\; f_{\operatorname{mag}}(\rvs).
\end{equation}
We couple this with an adaptive sparsity objective on the gate:
\begin{equation}
\label{eq:l1mask}
\mathcal{L}_{\ell_1}(\vm) \;=\; \bigl\| \operatorname{ReLU}\!\big(\mathrm{Gate}(\rvs)\big) \bigr\|_{1},
\end{equation}
where the target sparsity threshold is increased smoothly using the cosine schedule in \eqref{eq:mi-schedule}. 
This schedule provides fine-grained control over the active fraction of latent dimensions while preserving exploration efficiency.

\subsection{Details on Skill Library Learning}\label{app:misl}
We use DIAYN~\citep{eysenbach2018diversity} and METRA~\citep{park2024metra} as our skill library learning framework. 

For DIAYN, let $z\!\sim\!p_\eta(z)$ be a skill sampled per episode/segment and
$s\!\sim\!d^{\pi_\psi}(\cdot\mid z)$ the (discounted) state distribution
induced by the skill–conditioned policy $\pi_\psi(a\mid s,z)$.
Define
\begin{equation}
\mathcal{F}(\psi,\eta)
\;=\;
\underbrace{\mathcal{H}[A \mid S,Z]}_{\text{max-ent exploration under skills}}
\;-\;
\mathcal{H}[Z \mid S]
\;+\;
\mathcal{H}[Z],
\end{equation}

where entropies are taken under $z\!\sim\!p_\eta$, $s\!\sim\!d^{\pi_\psi}(\cdot\mid z)$,
and $a\!\sim\!\pi_\psi(\cdot\mid s,z)$.
Expanding the KL terms gives
\begin{equation}
\mathcal{F}(\psi,\eta)
=
\mathcal{H}[A \mid S,Z]
+
\mathbb{E}_{z\sim p_\eta,\, s\sim d^{\pi_\psi}(\cdot\mid z)}
\!\big[\log p(z\mid s)\big]
-
\mathbb{E}_{z\sim p_\eta}\!\big[\log p_\eta(z)\big].
\end{equation}
Using a variational classifier $q_\phi(z\mid s)$,
we obtain the tractable lower bound
\begin{equation}
\mathcal{F}(\psi,\eta)
\;\ge\;
\mathcal{G}(\psi,\eta,\phi)
\;\coloneqq\;
\mathbb{E}_{z\sim p_\eta,\, s\sim d^{\pi_\psi}(\cdot\mid z)}
\!\Big[
\underbrace{\mathcal{H}\!\big(\pi_\psi(\cdot\mid s,z)\big)}_{\text{entropy of }\pi}
+
\log q_\phi(z\mid s) - \log p_\eta(z)
\Big].
\end{equation}
For METRA, we consider the skill reward as:
\begin{equation}
r_\phi(\rvs,\rvz,s\rv')=\big(\phi(\rvs')-\phi(\rvs)\big)^\top \rvz .
\end{equation}
and the total objective is then as: 
\begin{equation}
\mathcal{L}_{\text{METRA}}(\phi,\lambda)=
\mathbb{E}_{(\rvs,\rvz,\rvs')\sim\mathcal{D}}
\!\Big[\big(\phi(\rvs')-\phi(\rvs)\big)^\top \rvz
+\lambda\,\mathrm{clip}_\varepsilon\!\big(1-\|\phi(\rvs')-\phi(\rvs)\|_2^2\big)\Big],\quad \lambda\!\ge\!0.
\end{equation}

Both are optimized using SAC~\citep{haarnoja2018soft} to learn the exploration policy. Hyperparameters are given in Section~\ref{sec:hp}.

\subsection{Details on Curriculum Learning}
We follow a UED-style scheduler for environment selection. At each step, we compute an ensemble-disagreement proxy $\Delta_T$ (model-error hardness) per environment
from its buffer:
\begin{equation}
\Delta_T \;=\; \mathbb{E}_{(s,a)\sim \mathcal{B}_T}
\!\left[\operatorname{Tr}\!\big(\mathrm{Cov}_{m=1}^{M}\,[\,f_{\theta^{(m)}}(s,a)\,]\big)\right].
\end{equation}
We then normalize scores before sampling:
\begin{equation}
\tilde{\Delta}_T \;=\; \frac{\Delta_T - \min_{T'} \Delta_{T'}}{\max_{T'} \Delta_{T'} - \min_{T'} \Delta_{T'} + \varepsilon}.
\end{equation}
A Boltzmann distribution with temperature $\eta$ emphasizes harder environments:
\begin{equation}
p_\eta(T) \;=\; \frac{\exp\!\big(\tilde{\Delta}_T / \eta\big)}{\sum_{T' \in \mathcal{T}} \exp\!\big(\tilde{\Delta}_{T'} / \eta\big)}.
\end{equation}
Then, we mix domain randomization with error-driven selection: with probability $p_{\mathrm{DR}}$ (or if any
$\mathcal{B}_T$ is empty), we sample environment parameters at random; otherwise, we sample according to
$p_\eta$:
\begin{equation}
u \sim \mathcal{U}[0,1],\qquad
T_{\text{next}} \sim
\begin{cases}
\mathrm{DomainRandomisation}(\Theta), & \text{if } u < p_{\mathrm{DR}} \ \text{or}\ \exists\,T:\ |\mathcal{B}_T|=0,\\[4pt]
\mathrm{Cat}\!\big(p_\eta(\cdot)\big), & \text{otherwise}.
\end{cases}
\end{equation}

{To compute the disagreement term, we use an ensemble of transition functions
$f_\theta(\rvs, \rva)$ corresponding to the RSSM dynamics in the Dreamer model, following~\citet{ball2020ready, sekar2020planning, mendonca2021discovering}. Concretely, we
train an ensemble of one-step predictors that map the current model state and action to the next
model state. The ensemble is optimized jointly with the world model on latent transitions
$\rvs' \sim p_\theta(\rvs' \mid \rvs, \rva)$, as defined in Eq.~\ref{eq:dreamer}.}

\subsection{{Expandable Latent Space}} \label{app:grow}
When transitioning from task $T_i$ to $T_{i+1}$, we allow the latent dimensionality to grow,
$d_{i+1} > d_i$. 
We follow a simple over-parameterized heuristic that is consistent with practice and then rely on structure learning to select the task-relevant subspace.
For each benchmark we initialize the Dreamer latent dimensionality $d_1$ to a fixed value and keep it shared across tasks at the beginning (e.g., $d_1 = 512$ for Meta-World, RoboSuite, and Kitchen, which have high-dimensional visual observations, and $d_1 = 256$ for DMControl). When moving from task $T_i$ to $T_{i+1}$, we allow the latent dimensionality to grow by a fixed increment, $d_{i+1} = d_i + \Delta d$. 
where we set $\Delta d = 32$ in all experiments. The choice of $d_1$ and $\Delta d$ is therefore a heuristic design decision rather than an oracle choice of the ``correct'' dimension. In practice, this slight over-parameterization is benign because our structure-learning module (through the sufficiency and minimality objectives and the learned mask $m^i$) identifies and uses only the task-relevant subset of dimensions (the MIST states), while irrelevant coordinates are effectively suppressed.

Concretely, we fine-tune the encoder and RSSM parameters up to the latent layer
for $T_i$ and adapt the final linear layer that maps the encoder/RSSM features into
$\mathbb{R}^{d_{i+1}}$. The additional $d_{i+1} - d_i$ coordinates
are initialized randomly and trained for $T_{i+1}$. The decoders (observation, reward, and
continuation) are similarly extended to read the expanded latent vector, but we fine-tune their
parameters for the first $d_i$ dimensions and learn the additional weights for the new
coordinates.
\subsection{{Additional Details on Representation Learning Components}}
\label{app:components}
\paragraph{Unstructured Latent States.}
The Dreamer world model produces latent states
$\rvs_t \in \mathbb{R}^{d}$ through the RSSM encoder
$q_\theta(\rvs_t \mid \rvo_t, h_t, \rva_{t-1})$.
These latents are \emph{unstructured}: no constraints are imposed on their
factorization, and all coordinates may encode arbitrary mixtures of task-relevant
and task-irrelevant features. During structure learning,
the encoder parameters generating $\rvs_t$ remain \textbf{frozen} so that all
subsequent structure-learning objectives operate over a fixed latent basis.

\paragraph{Structure Learning.}
Given the unstructured latents $\rvs_t$, we identify a structured subspace
$S^{\mathrm{min}}_i$ for each task $T_i$ via a learnable mask
$m^i \in (0,1)^d$ and two task-specific objectives (Eq.~\ref{eq:suffi} \& Eq.~\ref{eq:mini}). 
During this stage, the RSSM encoder and Dreamer dynamics are \textbf{frozen};
only the mask $m^i$ and the MI estimator parameters receive gradients.

\paragraph{MIST States.}
For each task $T_i$, the MIST representation is defined as
\[
\tilde{\rvs}^i_t = m^i \odot \rvs^i_t \in \S^{\mathrm{min}}_i,
\]
which retains only the coordinates necessary for reward prediction and
decision making. Policies and value functions $\pi_\psi(\rva_t\mid\tilde{\rvs}^i_t)$
operate exclusively on this subspace.

\paragraph{Active Probing Skills.}
When encountering a new task, the agent collects informative trajectories via
latent skills $\rvz \sim p_\eta(\rvz)$.
Actions follow a skill-conditioned policy
$\pi_\varepsilon(\rva_t \mid \rvs_t, \rvz)$, trained using the mutual-information
based MISL objective (Eq.~\ref{eq:misl}). 
Here only the skill policy $\eta$ and discriminator $q_\phi$ receive gradients;
the world model and encoders stay \textbf{fixed} during skill learning.

\paragraph{Segment Separability.}
To select informative skills, we measure how distinguishable two segments
generated from different latent-factor values are. Let
$\rvo^{(v)}_{t:t+k}$ and $\rvo^{(v')}_{t:t+k}$ be observation segments produced
under skill $\rvz$. After encoding them using $\phi_{\xi}$,
we apply an InfoNCE objective (Eq.~\ref{eq:contrastive}).
Gradients update the segment encoder $\phi$ and the 
skill parameters $\varepsilon$.

Table~\ref{tab:gradflow} summarizes which objectives update which components.

\begin{table}[h]
\centering
\begin{tabular}{lccc}
\hline
\textbf{Module} &
\textbf{WM Loss} &
\textbf{Structure Losses}  &
\textbf{Skill Losses}
\\
\hline
Dreamer encoder / RSSM      & \checkmark & frozen & frozen \\
Reward / obs decoders       & \checkmark & \checkmark & frozen \\
Mask $m^i$                  & frozen     & \checkmark & frozen \\
MI estimator (MINE)         & frozen     & \checkmark & frozen \\
Skill policy $\varepsilon$         & frozen     & frozen     & \checkmark \\
Skill discriminator $q_\phi$& frozen     & frozen     & \checkmark \\
Segment encoder $\phi_\xi$  & frozen     & frozen     & \checkmark \\
Policy / value $\psi$       & frozen     & frozen     & \checkmark (RL) \\
\hline
\end{tabular}
\caption{Summary of gradient flow across modules. A checkmark indicates parameters updated under the corresponding losses.}
\label{tab:gradflow}
\end{table}

\subsection{{Evaluation on the Learned Representations}} \label{app:rsquare}
As illustrated in Fig.~\ref{fig:res_1}, we assess representation quality by measuring how well the
learned latents predict the ground-truth simulator factors. Since the latent space can be
over-parameterized, a single factor may be encoded across multiple coordinates. We therefore first
fix all model parameters and only use the encoders to map observations to latent vectors
$\rvs_t^i$, so that no further learning happens in the world models during this evaluation.

Given the assignment of latent dimensions to each ground-truth variable, we group the corresponding
coordinates and attach a separate probe network to every group. Each probe is a two-layer MLP with
hidden size 128, trained to regress all ground-truth factors associated with that group, and we
report the coefficient of determination ($R^2$) between the predictions and the true values. 

To train the probes, we randomly split the evaluation rollouts into two parts: 40\% of the data is used to fit the MLPs and the remaining 60\% is held out for computing the correlation metrics. The
rollouts themselves are obtained by sampling random configurations of the underlying ground-truth latent factors. We use MSE loss for continuous variables and cross entropy loss for categorical variables. 

After training the MLPs, we use the coefficient of determination ($R^2$) to determine the "matching quality" of learned representation and the ground-truth ones. 
The $R^2$ coefficient quantifies how well the predicted values $\hat{x}_i$ explain the variance of a
ground-truth variable $x_i$. For continuous variables, it is defined as
\begin{equation}
R^2
= 1 - 
\frac{\sum_{i} (x_i - \hat{x}_i)^2}
     {\sum_{i} (x_i - \bar{x})^2},
\end{equation}
where $\bar{x} = \tfrac{1}{N}\sum_i x_i$ denotes the empirical mean of the ground-truth values. Similarly, for categorical variables, we can use the most frequent category in
the dataset to approximate $\bar{x}$, and then measure the distance by equal: $\delta_{x_i=\hat{x_i}}$.

\section{Details on Task Settings and Hyperparameters}

\subsection{Task Settings} 
We evaluate our framework across a diverse set of standard continuous control (locomotion) and robotic manipulation benchmarks. 
From DMControl~\citep{tunyasuvunakool2020dm_control}, we consider three representative locomotion tasks—\textit{Cheetah Run}, \textit{Walker}, and \textit{Reacher}. To assess manipulation and generalization, we further include Meta-World~\citep{yu2020meta}, a widely used multi-task benchmark covering diverse object-centric skills, and extend to more realistic robotic platforms such as Franka-Kitchen~\citep{gupta2019relay} and Robosuite~\citep{zhu2020robosuite}. 

\subsubsection{Benchmarks}
\paragraph{DMControl} The DeepMind Control Suite (DMControl)~\citep{tunyasuvunakool2020dm_control} is a standard continuous-control benchmark based on MuJoCo physics. We evaluate on three representative tasks:
\begin{itemize}
    \item Cheetah-Run: a planar cheetah agent optimized for speed, testing fast locomotion and stability.
    \item Walker: a bipedal agent balancing and walking forward, emphasizing coordination.
    \item Reacher: a 2-link arm reaching for randomly placed targets, measuring precise control and sample efficiency.
\end{itemize}
\begin{figure}[h]
\begin{center}
\centerline{\includegraphics[width=\columnwidth]{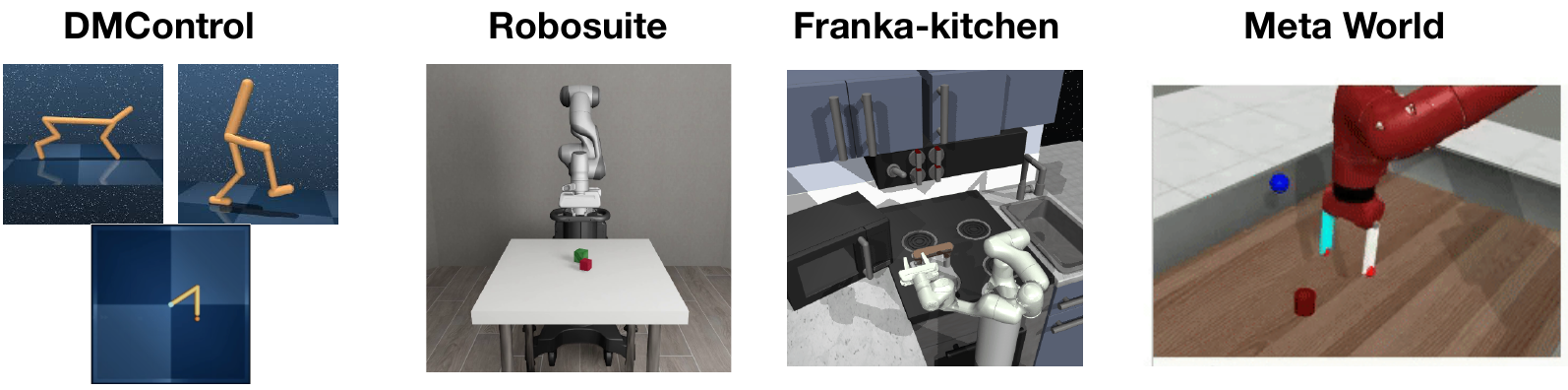}}
\caption{Visualization on tested benchmarks.}
\label{fig:vis}
\end{center}\vspace{-5mm}
\end{figure}
\paragraph{Meta-World} Meta-World~\citep{yu2020meta} is a large multi-task benchmark for robotic manipulation. Each task involves a 7-DoF robotic arm manipulating the same object but with opposite goals, providing an ideal testbed for multi-task interference. 

\paragraph{Franka-kitchen} The Franka-Kitchen benchmark~\citep{gupta2019relay} simulates a realistic Franka Emika Panda robot arm interacting with common kitchen appliances (microwave, lights, burners, cabinets). Tasks are composed from multiple sub-goals, requiring long-horizon planning, object interaction, and compositional skill reuse.

\paragraph{Robosuite} Robosuite~\citep{zhu2020robosuite} is a suite of MuJoCo-based robotic benchmarks designed for general-purpose robot learning. It provides a diverse set of manipulation tasks with realistic object dynamics and robot arms (including Sawyer and Panda).

\subsubsection{Single Task Learning}
We evaluate single-task learning across four suites. In \textbf{DMControl}, we use \textit{Cheetah Run}, \textit{Walker}, and \textit{Reacher}. In \textbf{Meta-World}, we use \textit{Box-Close}, \textit{Lever-Pull}, \textit{Assembly}, \textit{Bin-Picking}, \textit{Handle-Pull}, \textit{Door-Unlock}, \textit{Faucet-Close}, and \textit{Drawer-Open}. In \textbf{RoboSuite}, we use \textit{Lift} and \textit{Pick \& Place}. In \textbf{Franka Kitchen}, we use \textit{Microwave}, \textit{Kettle}, \textit{Stove}, \textit{Light}, and \textit{Cabinet}. 

We train for \(10^{6}\) environment steps on DM Control, \(5\times10^{6}\) on Meta-World, \(5\times10^{6}\) on RoboSuite, and \(5\times10^{4}\) on Franka Kitchen.

\subsubsection{Generalization tasks}
\textbf{Skill generalization.} 
For \textbf{RoboSuite}, we evaluate stacking tasks that can be composed from skills learned on \textit{Lift} and \textit{Pick \& Place}. 
For \textbf{Franka Kitchen}, we transfer previously learned skills to long-horizon sequences using the extended setup from \cite{chen2024scar}. 
We consider two multi-task sequences: 
\textbf{MT-1}: \textit{turn on microwave} $\rightarrow$ \textit{move kettle} $\rightarrow$ \textit{turn on stove} $\rightarrow$ \textit{turn on light}; 
\textbf{MT-2}: \textit{turn on microwave} $\rightarrow$ \textit{turn on stove} $\rightarrow$ \textit{turn on light} $\rightarrow$ \textit{slide cabinet to the right}. 
We use \(10^{6}\) adaptation steps for RoboSuite and \(1.25\times10^{5}\) for Franka Kitchen.

\textbf{Compositional \& unseen generalization.} 
For \textbf{Meta-World}, we test compositional generalization to \textit{door-open} and \textit{drawer-close}, and unseen-task generalization to \textit{coffee-push}, \textit{sweep-into}, \textit{push-wall}, \textit{reach-wall}, and \textit{plate-slide} (randomly selected). 
All adaptations use \(2.3\times10^{5}\) environment steps.

\begin{figure}[h]
\begin{center}
\centerline{\includegraphics[width=\columnwidth]{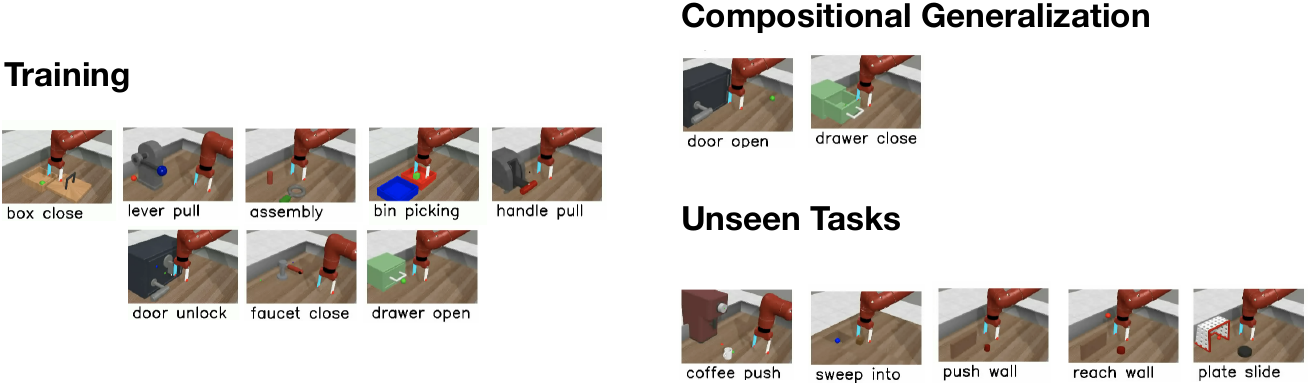}}
\caption{Visualization on the tasks used for compositional generalization and unseen tasks. }
\label{fig:vis_mw}
\end{center}\vspace{-5mm}
\end{figure}

\subsection{Environment Settings}\label{app:env_design}
\textbf{RoboSuite.}
We design controlled multi-object scenes to encourage recovery of system variables and enable active probings. 
Each episode initializes \emph{three} cubes, with two cubes sharing identical color and texture. 
We construct a bank of ten environment variants that systematically vary a single physical factor at a time—(i) mass (\texttt{body\_mass}), (ii) friction coefficients, and (iii) orientation components (e.g., Euler angles)—while keeping all other factors fixed. 
Within any episode, only \emph{two} of the three cubes differ along the currently controlled factor. 
This parametrization yields clear probing targets and facilitates learning an active probing policy.

\textbf{Meta-World \& Franka Kitchen.}
Given the more limited low-level control in \textit{Meta-World}, we adopt multi-object settings analogous to \textit{Franka Kitchen}, which naturally contains diverse, manipulable objects. 
For Meta-World, we reuse the \emph{Continual Bench} setup from \cite{liu2025continual} to explore environments featuring drawers, doors, boxes, bins, and handles, in order to learn object-centric latent representations before downstream tasks. 
All baselines are granted the same phase-one interaction protocol and budget to ensure fairness. 
Unless otherwise specified, we allocate  $10\%$ of the total environment steps for this representation-learning phase across suites.

\subsection{Hyperparameters}\label{sec:hp}
We build on the official \textit{Dreamer-v3} codebase (including \textit{DMControl}/MuJoCo tasks). 
For \textit{Meta-World}, we adopt the Dreamer-v3 hyperparameters as specified in the \textit{TD-MPC2} paper. 
For \textit{Franka Kitchen} and \textit{RoboSuite}, we use the same configuration as Meta-World due to task similarity. 
Baseline implementations follow their public releases wherever possible: \textit{DINO-WM} is based on its official learning setup, but we replace the original MPC component with either PPO or SAC to match our environments; \textit{TD-MPC2} is run with its original settings on \textit{DMControl}. 
Beyond these, our method introduces phase-specific hyperparameters for (i) the exploration policy, (ii) MIST learning, and (iii) adaptation. 
The specification for (i) - (iii) is provided in Table~\ref{tab:hparams1}-\ref{tab:hparams2}. 
The hyperparameters of SAC and PPO are specified in Table~\ref{tab:sac-hparams}-\ref{tab:ppo-hparams}.

\subsection{Details on Computational Requirements}
\label{app:cost}
All experiments are conducted on 4 NVIDIA A100 GPUs with 80GB memory. 
We report the training cost in GPU days and compare against the base model (Dreamer-v3). 
As shown in Table~\ref{tab:gpu_days}, our method requires slightly more compute while achieving better generalization.

\begin{table}[h]
\centering
\caption{Training cost (GPU days) comparison with Dreamer-v3 on different suites.}
\label{tab:gpu_days}
\begin{tabular}{lcc}
\toprule
Task Suite & Ours & Dreamer-v3 \\
\midrule
DM Control      & 0.35 & 0.25 \\
Meta-World      & 0.75 & 0.50 \\
Franka Kitchen  & 0.35 & 0.30 \\
RoboSuite       & 0.75 & 0.60 \\
\bottomrule
\end{tabular}
\end{table}

\begin{table}[t]
\centering
\caption{Agent probing exploration and world model hyperparameters for \mist.}
\label{tab:hparams1}
\small
\begin{tabularx}{\columnwidth}{@{}L{0.63\columnwidth} C{0.32\columnwidth}@{}}
\toprule
\textbf{Hyperparameter} & \textbf{Value} \\
\midrule
\multicolumn{2}{@{}l}{\textbf{General}} \\
Batch size & 128 (DMControl); 512 (Others) \\
Ratio of imagined samples & 0.5 (DMControl); 0.2 (Others) \\
Replay buffer size & $10^6$ (DMControl); $10^5$ (Others) \\
\midrule
\multicolumn{2}{@{}l}{\textbf{Active Probing}} \\
Learning rate & 0.0001 \\
Replay buffer size & $10^6$ (DMControl); $10^5$ (Others) \\
Contrastive encoder layers & 3 \\
Contrastive hidden units & 256 \\
Episodes per batch & 16 (DMControl); 32 (Others) \\
Discount factor $\gamma$ & 0.99 \\
Policy network layers & 2 \\
Policy network hidden units & 1024 \\
Gradients per episode & 50 (DMControl); 100 (Others) \\
MISL algorithm & DIAYN (DMControl); METRA (Others) \\
Target smoothing coef. & 0.995 \\
Entropy coef. & 0.01 \\
\midrule
\multicolumn{2}{@{}l}{\textbf{World Model}} \\
Dreamer deterministic size & 512 for Meta World, 256 for Others \\
Dreamer stochastic size & 16 \\
Reward prediction layers & 2 \\
Reward pred. hidden units & 256 (DMControl); 512 (Others) \\
Policy network layers & 3 \\
Policy net hidden units & 256 (DMControl); 512 (Others) \\
Minimal score coef. & $10^{-3}$ (DMControl); $10^{-2}$ (Others) \\
Sufficiency score coef. & $10^{-3}$ (DMControl); $10^{-2}$ (Others) \\
\bottomrule
\end{tabularx}
\end{table}

\begin{table}[t]
\centering
\caption{Training, adaptation, and curriculum hyperparameters for \mist.}
\label{tab:hparams2}
\small
\begin{tabularx}{\columnwidth}{@{}L{0.63\columnwidth} C{0.32\columnwidth}@{}}
\toprule
\textbf{Hyperparameter} & \textbf{Value} \\
\midrule
\multicolumn{2}{@{}l}{\textbf{Training}} \\
Training steps & $5{\times}10^{6}$ (MW, RS); $5{\times}10^{4}$ (Kitchen); $10^{6}$ (DMControl) \\
Maximize MINE after & $10\%$ of total steps \\
\midrule
\multicolumn{2}{@{}l}{\textbf{Adaptation}} \\
Adaptation steps & $2.3{\times}10^{5}$ (Meta-World) \\
Sparsity & 0.35 \\
Maximize MINE (start) & 0 \\
\bottomrule
\end{tabularx}
\end{table}

\begin{table}[t]
\centering
\begin{tabular}{ll}
\hline
\textbf{Hyperparameter} & \textbf{Value} \\
\hline
Batch size                         & $128$ for Robsouite, $256$ for Meta-World \\
Network architecture               & MLPs \\
Actor / critic size                & 3 fully connected layers with 256 units \\
Non-linearity                      & ReLU \\
Policy initialization              & Standard Gaussian \\
Policy learning rate               & $3 \times 10^{-4}$ \\
Optimizer                          & Adam \\
Adam $\beta$ (policy)              & $(0.9, 0.999)$ \\
Adam $\beta$ (Q-function)          & $(0.9, 0.999)$ \\
Discount                           & $0.99$ \\
\hline
\end{tabular}
\caption{{SAC hyperparameters for Meta-World and RoboSuite.}}
\label{tab:sac-hparams}
\end{table}

\begin{table}[t]
\centering
\begin{tabular}{ll}
\hline
\textbf{Parameter} & \textbf{Value} \\
\hline
Network architecture               & MLPs \\
Actor / critic size                & $3 \times 256$ units for DMC, $3 \times 400$ units for Kitchen \\
Non-linearity                      & ReLU \\
Observation normalization   & Yes \\
Reward normalization        & Yes \\
Reward clipping (stddev.)   & 10 \\
Batch size              & 128 \\
Policy trust region         & 0.2 \\
Value trust region          & No \\
Advantage normalization     & Yes \\
Entropy penalty scale       & 0.01 \\
Discount factor             & 0.997 \\
GAE $\lambda$               & 0.95 \\
Learning rate               & $3 \times 10^{-4}$ \\
Gradient clipping           & 0.5 \\
Adam $\varepsilon$          & $10^{-5}$ \\
\hline
\end{tabular}
\caption{{PPO hyperparameters for DMC and Kitchen.}}
\label{tab:ppo-hparams}
\end{table}

\begin{table}[t]
\centering
\caption{{Curriculum learning hyperparameters.}}
\begin{tabular}{lc}
\toprule
\textbf{Hyperparameter} & \textbf{Value} \\
\midrule
Temperature  $\eta$     & $0.1$ \\
Domain randomization prob.  $p_{\mathrm{DR}}$  & $0.3$ \\
Normalization constant  $\varepsilon$    & $10^{-6}$ \\
Ensemble size  $M$      & $5$ \\
\bottomrule
\end{tabular}
\label{tab:curri_hparams}
\end{table}

\section{Full Results}
\subsection{Full Results on Representation Probing}
\begin{table}
\centering
\captionof{table}{\textbf{Probing results ($R^2$) under different settings}. 
The pink region corresponds to varying the exploration data using different collection methods, 
while the green region corresponds to varying the world-modeling methods while keeping the data fixed. The average is computed across all factors excluding the size.}
\label{tab:representation}
{\small
\setlength{\tabcolsep}{6pt}
\renewcommand{\arraystretch}{1.15}
\begin{tabular}{lc|cccccccc}
\toprule
 \multicolumn{2}{c}{Methods} & Size  &  C-pos & C-Orient & G-pos & G-orient & G-openness & C-Mass & Average\\
\midrule
\midrule
\rowcolor{panelA}
\multirow{4}{*}{\rotatebox[origin=c]{0}}
  & Random              & 0.01 & 0.92& 0.16& 0.48& 0.21& 0.06& 0.03 & 0.31\\
\rowcolor{panelA}
   & METRA           & 0.03 & 0.95& 0.85& 0.51& 0.49& 0.20& 0.55 & 0.59 \\
\rowcolor{panelA}
   & Curiosity Replay   & 0.02 & 0.96& 0.53& 0.23& 0.36& 0.15& 0.39 & 0.44\\
\rowcolor{panelA}
   & Plan2Explore      & 0.01 & 0.93& 0.59& 0.41& 0.26& 0.10& 0.40 & 0.45\\
\midrule
\rowcolor{panelB}
\multirow{3}{*}{\rotatebox[origin=c]{0}}
  & DINO-WM             & \textbf{0.76} & 0.75& 0.09& 0.53& 0.42& 0.00& 0.61 & 0.40\\
\rowcolor{panelB}
  & I-Factor          & 0.00 & 0.79& 0.41& 0.56& 0.47& 0.08& 0.59 & 0.48 \\
\rowcolor{panelB}
  & Denoised MDP        & 0.00 & 0.83& 0.30& 0.72& 0.65& 0.16& 0.68 & 0.56\\
\midrule
 & \mist               & 0.02 & \textbf{1.00}& \textbf{0.94}& \textbf{0.86}& \textbf{0.69}& \textbf{0.21}& \textbf{0.96} & \textbf{0.78}\\
\bottomrule
\end{tabular}
}
\end{table}
Table~\ref{tab:representation} reports the full representation–probing results; the averaged scores appear in Fig.~\ref{fig:res_1}(c) of the main paper. The trends mirror our main findings. \mist achieves the best overall performance. In the yellow block (exploration skills), METRA is consistently second-best. In the green block (world models), factored approaches (\textit{I-Factor} and \textit{Denoised MDP}) outperform DINO-WM on average. Notably, DINO-WM excels on the \emph{size} factor, likely due to its strong pretrained visual prior. 

\subsection{Full Results on Single Task Learning}
\begin{figure}[h]
\begin{center}
\centerline{\includegraphics[width=\columnwidth]{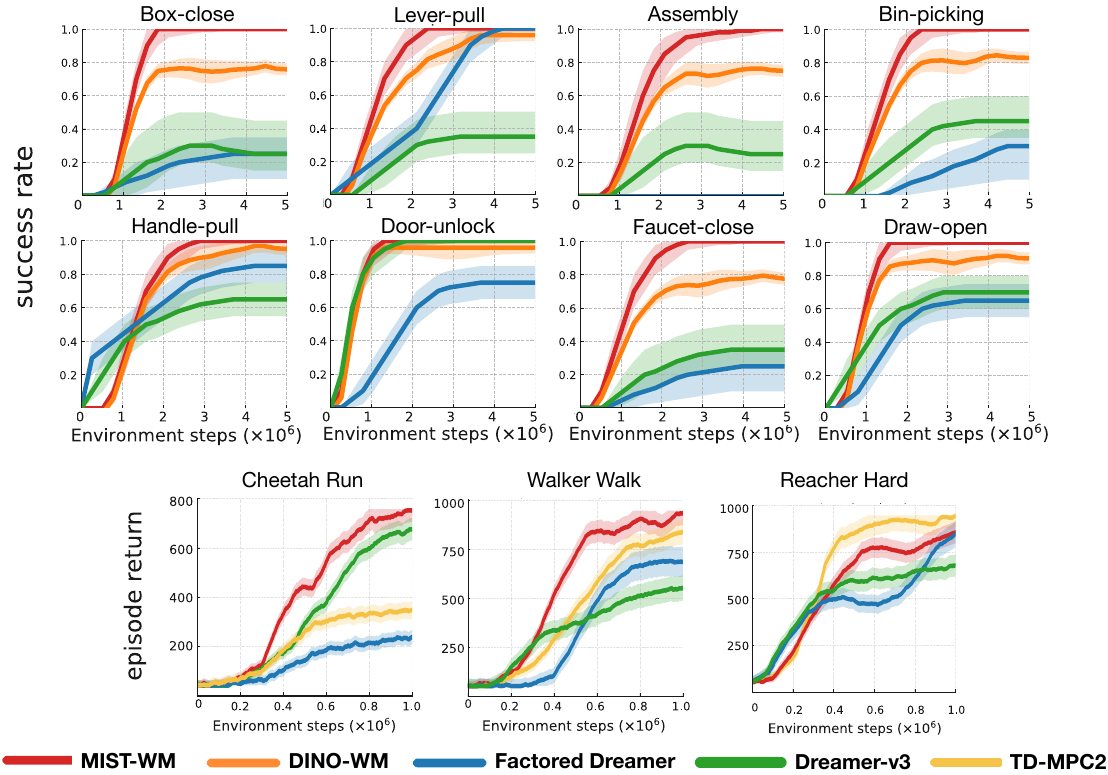}}
\caption{Results (across 5 seeds) on single-task learning, including \textbf{robotic manipulation} tasks (Meta World) and \textbf{locomotion} (DMControl).}
\label{fig:res_single_task}
\end{center}\vspace{-5mm}
\end{figure}

Fig.~\ref{fig:res_single_task} presents the complete single-task learning results, covering 8 tasks from Meta-World and 3 tasks from DMControl, complementing the summary shown in Fig.~\ref{fig:res_single_task_short}.

\begin{table}[t]
\centering
\caption{{Success rate at 2M environment steps (mean $\pm$ std over 5 seeds).}}
\label{tab:mt_2m}
\begin{tabular}{lccccc}
\toprule
Environment & Ours & DINO-WM & Factored Dreamer & Dreamer-v3 & TD-MPC2 \\
\midrule
Box-close & $\mathbf{0.94}\text{\scriptsize\color{gray}{ $\pm$ 0.02}}$ & 
            $0.75\text{\scriptsize\color{gray}{ $\pm$ 0.05}}$ &
            $0.15\text{\scriptsize\color{gray}{ $\pm$ 0.06}}$ &
            $0.26\text{\scriptsize\color{gray}{ $\pm$ 0.13}}$ &
            $0.09\text{\scriptsize\color{gray}{ $\pm$ 0.02}}$ \\

Lever-pull & $\mathbf{0.91}\text{\scriptsize\color{gray}{ $\pm$ 0.05}}$ &
             $0.72\text{\scriptsize\color{gray}{ $\pm$ 0.04}}$ &
             $0.38\text{\scriptsize\color{gray}{ $\pm$ 0.08}}$ &
             $0.29\text{\scriptsize\color{gray}{ $\pm$ 0.11}}$ &
             $0.06\text{\scriptsize\color{gray}{ $\pm$ 0.03}}$ \\

Assembly & $\mathbf{0.79}\text{\scriptsize\color{gray}{ $\pm$ 0.12}}$ &
            $0.61\text{\scriptsize\color{gray}{ $\pm$ 0.03}}$ &
            $0.06\text{\scriptsize\color{gray}{ $\pm$ 0.02}}$ &
            $0.22\text{\scriptsize\color{gray}{ $\pm$ 0.08}}$ &
            $0.12\text{\scriptsize\color{gray}{ $\pm$ 0.06}}$ \\

Bin-picking & $\mathbf{0.88}\text{\scriptsize\color{gray}{ $\pm$ 0.05}}$ &
              $0.71\text{\scriptsize\color{gray}{ $\pm$ 0.04}}$ &
              $0.10\text{\scriptsize\color{gray}{ $\pm$ 0.06}}$ &
              $0.31\text{\scriptsize\color{gray}{ $\pm$ 0.12}}$ &
              $0.08\text{\scriptsize\color{gray}{ $\pm$ 0.03}}$ \\

Handle-pull & $\mathbf{0.84}\text{\scriptsize\color{gray}{ $\pm$ 0.06}}$ &
               $0.80\text{\scriptsize\color{gray}{ $\pm$ 0.02}}$ &
               $0.61\text{\scriptsize\color{gray}{ $\pm$ 0.09}}$ &
               $0.53\text{\scriptsize\color{gray}{ $\pm$ 0.08}}$ &
               $0.15\text{\scriptsize\color{gray}{ $\pm$ 0.04}}$ \\

Door-unlock & $0.96\text{\scriptsize\color{gray}{ $\pm$ 0.02}}$ &
               $0.91\text{\scriptsize\color{gray}{ $\pm$ 0.04}}$ &
               $0.55\text{\scriptsize\color{gray}{ $\pm$ 0.08}}$ &
               $\mathbf{0.98}\text{\scriptsize\color{gray}{ $\pm$ 0.05}}$ &
               $0.13\text{\scriptsize\color{gray}{ $\pm$ 0.02}}$ \\

Faucet-close & $\mathbf{0.91}\text{\scriptsize\color{gray}{ $\pm$ 0.06}}$ &
                $0.68\text{\scriptsize\color{gray}{ $\pm$ 0.08}}$ &
                $0.23\text{\scriptsize\color{gray}{ $\pm$ 0.12}}$ &
                $0.18\text{\scriptsize\color{gray}{ $\pm$ 0.11}}$ &
                $0.10\text{\scriptsize\color{gray}{ $\pm$ 0.04}}$ \\

Draw-open & $\mathbf{0.97}\text{\scriptsize\color{gray}{ $\pm$ 0.02}}$ &
             $0.86\text{\scriptsize\color{gray}{ $\pm$ 0.04}}$ &
             $0.62\text{\scriptsize\color{gray}{ $\pm$ 0.09}}$ &
             $0.54\text{\scriptsize\color{gray}{ $\pm$ 0.08}}$ &
             $0.17\text{\scriptsize\color{gray}{ $\pm$ 0.05}}$ \\
\bottomrule
\end{tabular}
\end{table}

\begin{table}[t]
\centering
\caption{{DMControl performance at 1M steps (mean $\pm$ std).}}
\label{tab:dmc_1m}
\begin{tabular}{lccccc}
\toprule
Environment & Ours & DINO-WM & Factored Dreamer & Dreamer-v3 & TD-MPC2 \\
\midrule
Cheetah-Run & 
$\mathbf{731.6}\text{\scriptsize\color{gray}{ $\pm$ 32.9}}$ &
$165.2\text{\scriptsize\color{gray}{ $\pm$ 41.8}}$ &
$217.5\text{\scriptsize\color{gray}{ $\pm$ 16.4}}$ &
$661.0\text{\scriptsize\color{gray}{ $\pm$ 42.3}}$ &
$339.8\text{\scriptsize\color{gray}{ $\pm$ 31.6}}$ \\

Walker-Walk & 
$\mathbf{904.2}\text{\scriptsize\color{gray}{ $\pm$ 42.5}}$ &
$286.5\text{\scriptsize\color{gray}{ $\pm$ 37.4}}$ &
$557.2\text{\scriptsize\color{gray}{ $\pm$ 56.4}}$ &
$698.2\text{\scriptsize\color{gray}{ $\pm$ 52.5}}$ &
$867.5\text{\scriptsize\color{gray}{ $\pm$ 51.2}}$ \\

Reacher-Hard & 
$871.6\text{\scriptsize\color{gray}{ $\pm$ 38.5}}$ &
$346.2\text{\scriptsize\color{gray}{ $\pm$ 29.8}}$ &
$869.3\text{\scriptsize\color{gray}{ $\pm$ 33.9}}$ &
$702.4\text{\scriptsize\color{gray}{ $\pm$ 43.3}}$ &
$\mathbf{915.6}\text{\scriptsize\color{gray}{ $\pm$ 23.3}}$ \\
\bottomrule
\end{tabular}
\end{table}

\begin{figure}[t]
\begin{center}
\centerline{\includegraphics[width=\columnwidth]{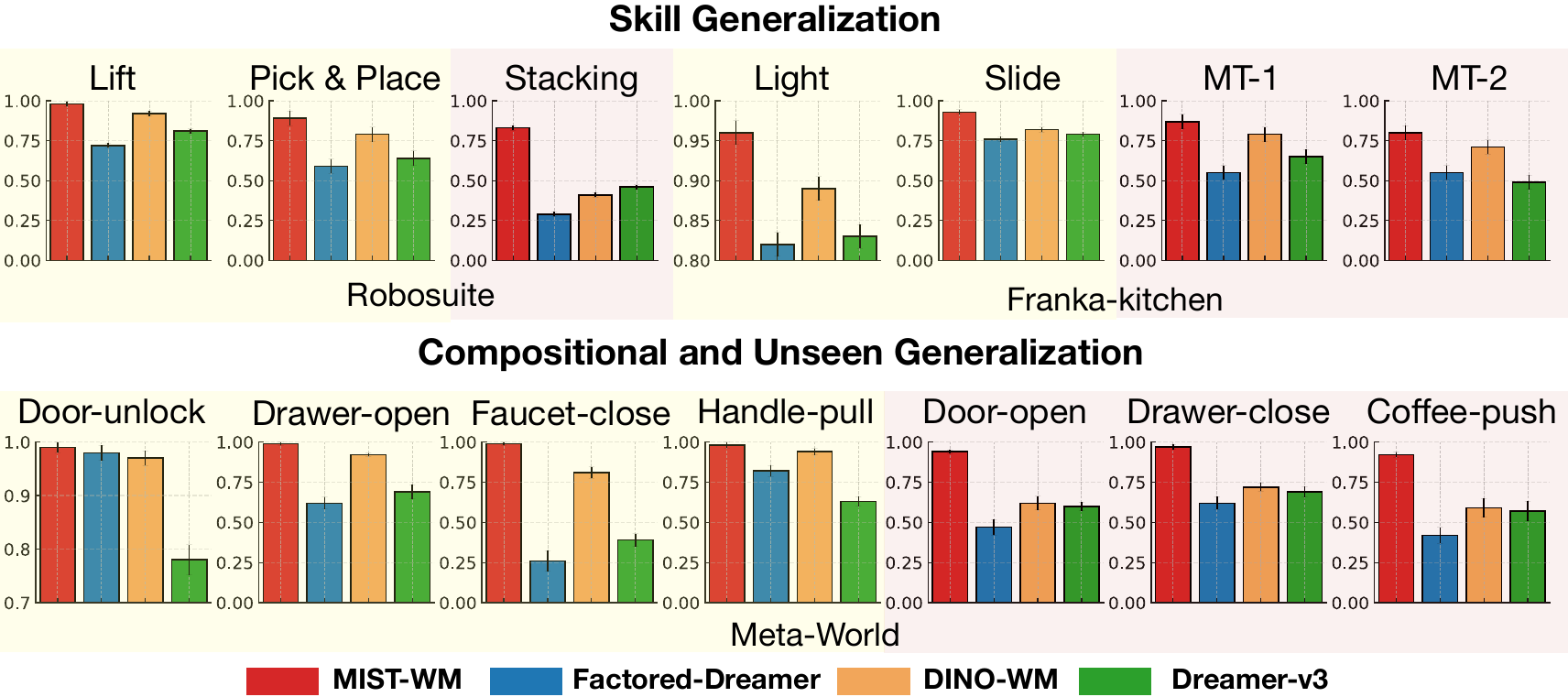}}
\caption{Success rate (across 5 seeds) on skill and compositional generalization, including tasks in Robosuite, Franka-kitchen, and Meta-World. Results in {light yellow} boxes are those on source tasks, and {light pink} are those on generalization tasks. {Error bars indicate the standard deviation.}}
\label{fig:res_gen}\vspace{-2mm}
\end{center}\vspace{-5mm}
\end{figure}

\subsection{Full Results on Generalization Tasks}
\begin{figure}[ht]
\begin{center}
\centerline{\includegraphics[width=\columnwidth]{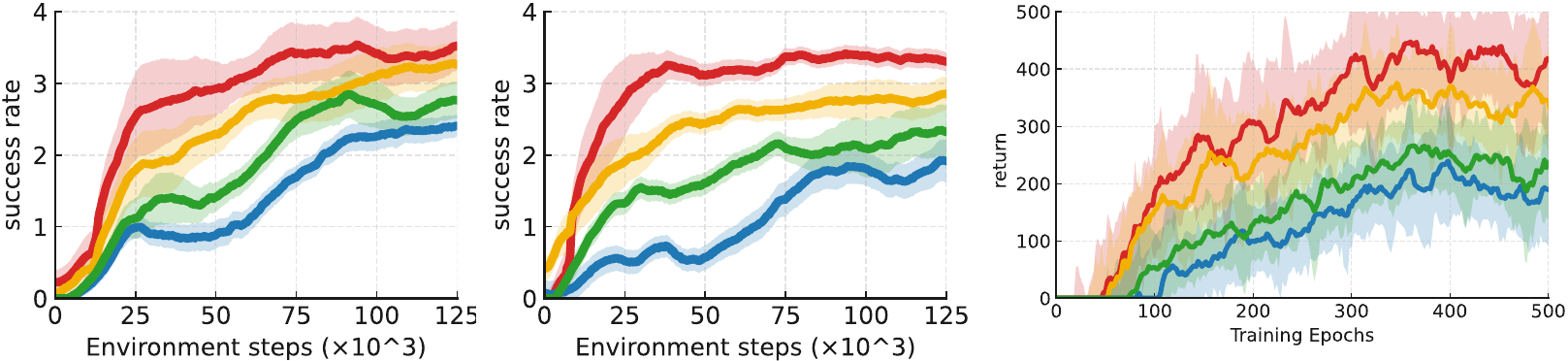}}
\caption{Results (across 5 seeds) on skill generalization tasks, where the left two figures are the skill generalization curves on Franka Kitchen MT-1 and MT-2; and the right figure is the generalization results on robosuite-stacking. The yellow curve indicates the results using DINO-WM.}
\label{fig:res_unseen_tasks}
\end{center}\vspace{-5mm}
\end{figure}

Figs.~\ref{fig:res_unseen_tasks} shows the complete results on skill generalization and on compositional/unseen tasks, respectively. The summary results are given in Fig.~\ref{fig:res_gen}. 

\subsection{Full Results on Ablation Studies}
\begin{figure}[h]
\begin{center}
\centerline{\includegraphics[width=\columnwidth]{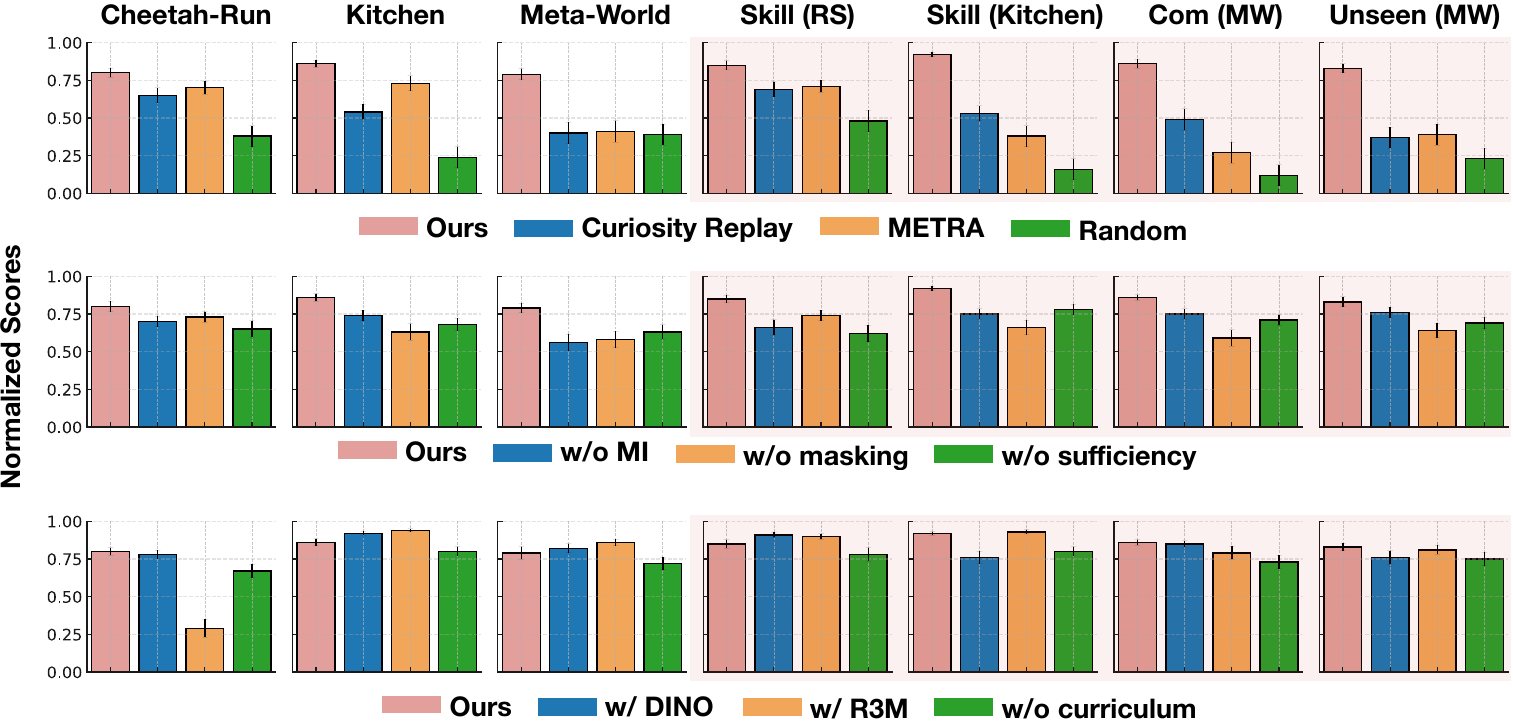}}
\caption{\textbf{Ablation studies.}
\textbf{Row 1—Exploration policy:} compare our skill-conditioned exploration against \emph{Random}, \emph{Intrinsic motivation} (Curiosity Replay), and \emph{MISL} (METRA).
\textbf{Row 2—Structure-learning terms:} remove the MI term in the \emph{minimality} score (w/o MI), remove \emph{masking} in the minimality score (w/o Mask), and remove the \emph{sufficiency} term (w/o Suff).
\textbf{Row 3—Backbones and curriculum:} replace our backbone with \emph{DINO-WM} or \emph{R3M}, and evaluate w/o curriculum. {Error bars indicate the standard deviation.}}
\label{fig:res_ablation}
\end{center}\vspace{-5mm}
\end{figure}

Fig.~\ref{fig:res_ablation} gives the full results on ablation studies, and the summary results are given in Fig.~\ref{fig:res_ablation_short} in the main paper.
\paragraph{Discussion: Choice of Skill-Learning Baselines.}
In the ablation studies, our primary goal is to examine how replacing our probing skill module with different unsupervised skill-learning
methods affects overall performance. While the original experiments used
METRA as a representative baseline, we have additionally included DIAYN
following the reviewer's suggestion. Table~\ref{tab:diayn_ablation}
summarizes the performance across three representative settings
(single-task, multi-task, and compositional generalization). We find that
METRA generally outperforms DIAYN, both as a standalone skill-learning
algorithm and when used as the skill-library initializer for MIST-WM.
However, we emphasize that the choice between METRA and DIAYN is not
central to our framework: both are only used to initialize the skill
library, while the probing skills are subsequently refined using our
separability objective in Eq.~\ref{eq:contrastive}. As a result, MIST-WM remains largely
agnostic to the choice of initial skill-learning algorithm.

\begin{table}[t]
\centering
\caption{Ablation comparing METRA and DIAYN as skill-library initializers.}
\label{tab:diayn_ablation}
\begin{tabular}{lccccc}
\toprule
Environment & Ours (METRA) & Ours (DIAYN) & METRA & DIAYN & Curious Replay \\
\midrule
Cheetah        & $\mathbf{741.0 \pm 25.6}$ & $731.6 \pm 32.9$ & $680.8 \pm 32.3$ & $664.5 \pm 39.6$ & $635.8 \pm 46.8$ \\
Box-close      & $\mathbf{0.94 \pm 0.02}$ & $0.79 \pm 0.04$ & $0.81 \pm 0.06$ & $0.59 \pm 0.10$ & $0.48 \pm 0.12$ \\
Coffee-push    & $\mathbf{0.89 \pm 0.05}$ & $0.82 \pm 0.03$ & $0.76 \pm 0.07$ & $0.62 \pm 0.05$ & $0.30 \pm 0.16$ \\
\bottomrule
\end{tabular}
\end{table}

\end{document}